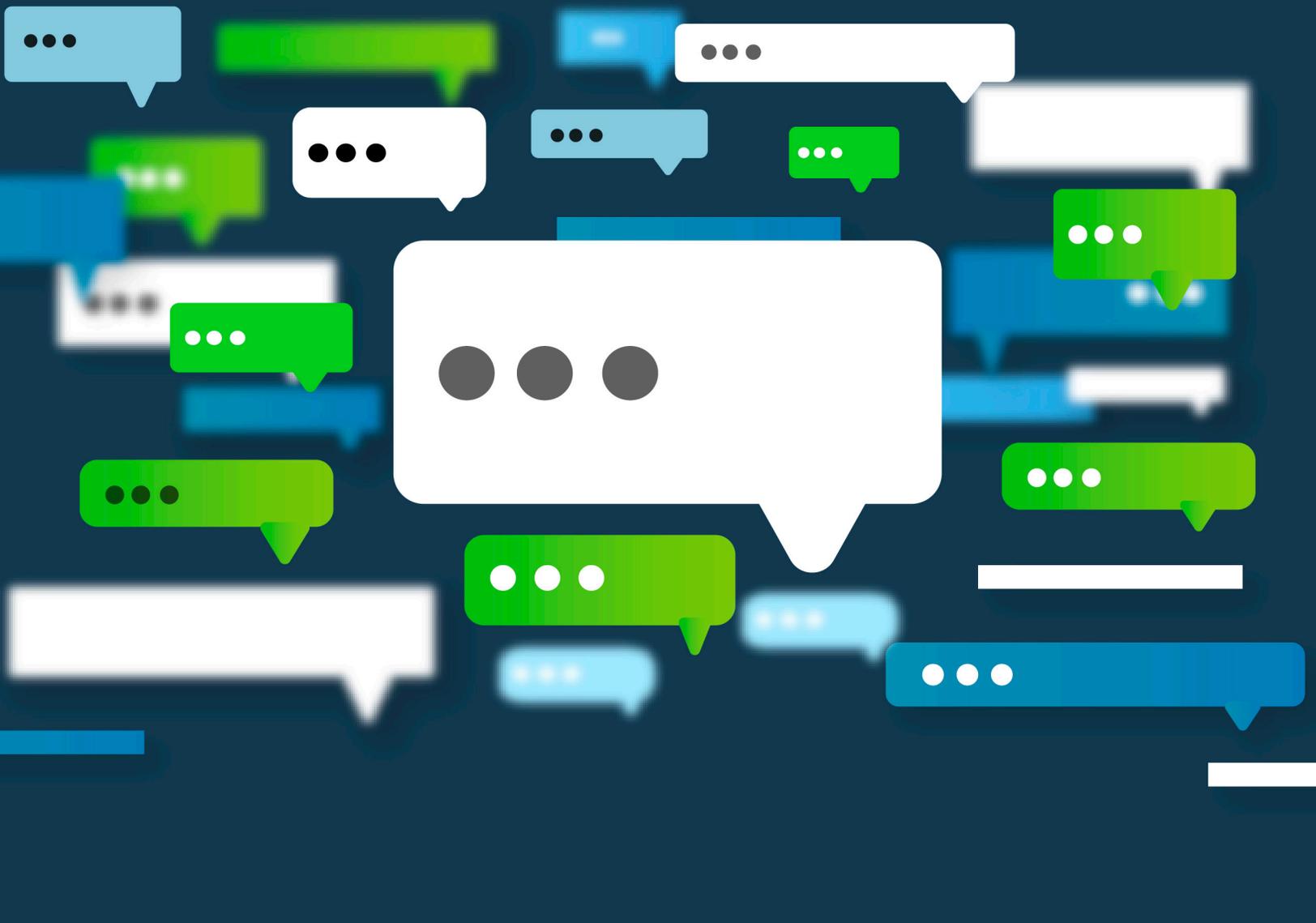

# A Fourth Wave of Open Data?
## Exploring the Spectrum of Scenarios for
# Open Data and Generative AI

Hannah Chafetz, Sampriti Saxena, and Stefaan G. Verhulst
May 2024

**THEGOVLAB**

**Open.Data.Policy.Lab**

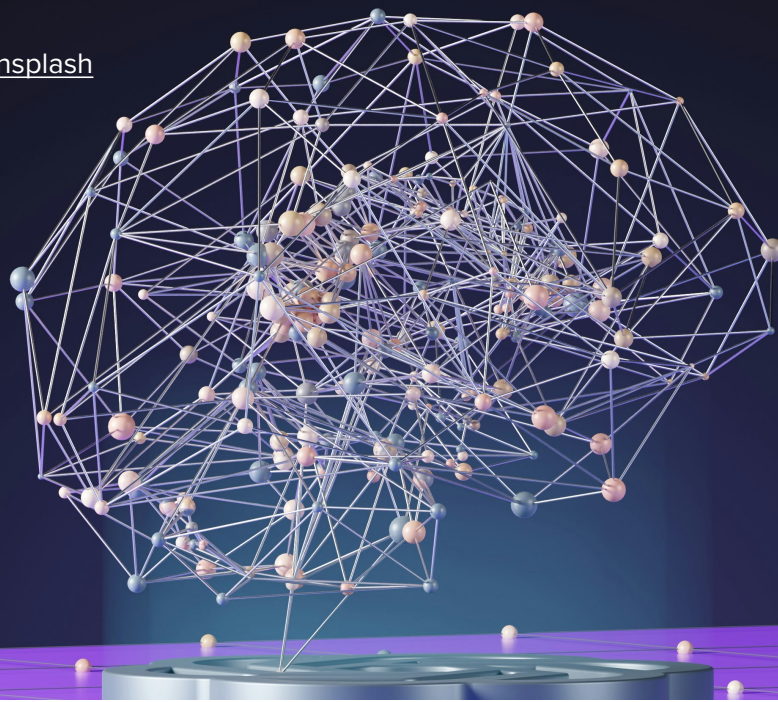

Photo by Growtika on Unsplash

## ABOUT THE AUTHORS

Hannah Chafetz and Sampriti Saxena are Research Fellows at The GovLab. Stefaan G. Verhulst is Co-Founder of The GovLab (New York) and The Data Tank (Brussels), and Research Professor at New York University.

## ACKNOWLEDGEMENTS

This work would not have been possible without the support and guidance of many individuals. In particular, we are grateful to Gretchen Deo and Sonia Cooper at Microsoft who helped us arrive at our initial topic of study and who were instrumental in securing the resources to make this paper possible. We would like to thank the following individuals for their generosity in sharing their time and valuable insights with us: Amra Dorjbayar, Ashley Farley, Bertrand Loison, Michael Tjalve, Oliver Wise, R.V. Guha, and William Hoffman.

Thank you to our peer reviewers for taking the time to review our first draft: Aimee Whitecroft, Alek Tarkowski, Amra Dorjbayar, Elena A. Kalogeropoulos, Elena Simperl, Gretchen Deo, Johanna Walker, Jorge Alvarado, Michael Tjalve, Oliver Bjornsson, Oliver Wise, Ramanathan V. Guha, Sonia Cooper, and William Hoffman. We are also grateful to our colleagues at The GovLab, especially Adrienne Schmoeker and Andrew J. Zahuranec, for their review and input.

Finally, we would like to thank all of the participants of our Open Data Action Labs for their engaged participation and contributions to this paper. The full list of participants can be found in Appendix 1.

# TABLE OF CONTENTS

# LIST OF FIGURES

# GLOSSARY OF TERMS

**Artificial Intelligence:** "is a machine-based system, that for explicit or implicit objectives infers, from the input it receives, how to generate outputs such as predictions, content, recommendations, or decisions."[1]

**Balanced Class Distribution:** When the training datasets for supervised learning techniques reflect the range of topics to be included in the model. Having a balanced class distribution ensures that the resulting output does not overfit one topic over another.

**Generative AI:** includes a range of models that can generate content based on training data. Generative Pretrained Transformers, or GPTs, for example, enable computers to perform tasks that mimic certain human brain functions such as interpreting language or generating code.

- **Generative AI Technologies/Applications:** "a type of artificial intelligence technology that can produce various types of content, including text, imagery, audio and synthetic data."[2]

- **Generative Pretrained Transformers (GPTs):** "a language model relying on deep learning that can generate human-like texts based on a given text-based input."[3]
  - "'Generative' means that it can create new data, [...] in the likeness of its training data. 'Pre-trained' means that the model has already been optimized based on this data, meaning that it does not need to check back against its original training data every time it is prompted. And 'Transformer' is a powerful type of neural network algorithm that is especially good at learning relationships between long strings of data, for instance sentences and paragraphs."[4]

---

**Hallucinations:** Generative AI models are trained to hallucinate outputs based on a set of prompts. However, sometimes the model can hallucinate outputs that are "incorrect or misleading [...]. These errors can be caused by a variety of factors, including insufficient training data, incorrect assumptions made by the model, or biases in the data used to train the model."[5] In this paper, we refer to hallucinations as these incorrect/low quality outputs.

**Knowledge Graph:** "Knowledge graphs provide explicit representations of knowledge, including descriptions of entities in terms of statements about them, relationships among entities and a variety of other conceptual structures, such as arguments, explanations and plans. Knowledge graphs include *ontologies*, i.e. representations that describe types of things in the world (entities) and facts about those entities."[6]

**Large Language Models (LLMs):** "deep learning algorithms that can recognize, summarize, translate, predict, and generate content using very large datasets."[7]

- **Deep Learning:** "a method used in developing AI systems which involves processing data in ways inspired by how the human brain works."[8]

- **Small Language Models:** "require less time and resources to maintain and can be operated inside a company's existing security perimeter."[9]

**Machine Learning:** "a subfield of artificial intelligence that gives computers the ability to learn without explicitly being programmed."[10]

---

- **Machine Learning Algorithm:** "a set of rules or processes used by an AI system to conduct tasks—most often to discover new data insights and patterns, or to predict output values from a given set of input variables."[11]

- **Machine Learning Model:** "a file that has been trained to recognize certain types of patterns. You train a model over a set of data, providing it an algorithm that it can use to reason over and learn from those data."[12]

**Natural Language Processing (NLP):** "the branch of AI focused on how computers can process language like humans do."[13]

**Open Data:** "publicly available data that can be universally and readily accessed, used and redistributed free of charge. It is structured for usability and computability."[14]

- **Open Government Data:** "promotes transparency, accountability and value creation by making government data available to all."[15] Public sector or government data holders often possess highly impactful datasets, which if opened up for public use, could make progress towards addressing public problems. These can include data from public surveys, like a census, or research and data collected by public organizations, like much of the work done by national and sub-national statistical organizations.

- **Open Research Data:** "refers to the data underpinning scientific research results that has no restrictions on its access, enabling anyone to access it."[16]

**Prompt Engineering:** "Prompt engineering is the practice of designing inputs for AI tools that will produce optimal outputs."[17]

**Retrieval Augmented Generation (RAG) Architectures:** "Retrieval Augmented Generation (RAG) is an architecture that augments the capabilities of a Large

---

Language Model (LLM) like ChatGPT by adding an information retrieval system that provides grounding data. Adding an information retrieval system gives you control over grounding data used by an LLM when it formulates a response."[18]

**Schemas:** "The structure or design of a database or database object, such as a table, view, index, stored procedure, or trigger. In a relational database, the schema defines the tables, the fields in each table, the relationships between fields and tables, and the grouping of objects within the database. Schemas are generally documented in a data dictionary. A database schema provides a logical classification of database objects."[19]

**Synthetic Data:** "Gartner defines synthetic data as a class of data that is artificially generated, as opposed to being collected via real-world observation. This synthetic data may be designed to mimic certain properties of real data or may be designed to simulate never-before-seen scenarios. It can also be used to make real data more representative by supplementing them with more diverse data."[20]

**Tabular Data:** "commonly known as structured data, refers to data organized into rows and columns, where each column represents a specific feature."[21]

**Third Wave of Open Data:** In the Third Wave of Open Data, data holders and users worked to break down data silos to open up data for the public good.[22] In this phase, activities are broadly clustered into four pillars: publishing with purpose; fostering partnerships and data collaboration; advancing open data at the subnational level; and prioritizing data responsibility and data rights.[23]

---

[18] Microsoft Learn. "Retrieval Augmented Generation (RAG) in Azure AI Search," November 20, 2023. https://learn.microsoft.com/en-us/azure/search/retrieval-augmented-generation-overview.

[19] esri. "Schema." Accessed March 26, 2024. https://support.esri.com/en-us/gis-dictionary/schema.

[20] Mueller, Henrike, Leo Gosland, Owen Courtney, Pavle Avramovic, and Valerie Marshall. "Can Synthetic Data Enable Data Sharing in Financial Services?" OECD AI Policy Observatory, May 25, 2023. https://oecd.ai/en/wonk/synthetic-data-financial-services.

[21] Xi Fang et al., "Large Language Models(LLMs) on Tabular Data: Prediction, Generation, and Understanding -- A Survey," *arXiv*, February 27, 2024, https://doi.org/10.48550/arXiv.2402.17944.

[22] Verhulst, Stefaan G. et al., *The Emergence of a Third Wave of Open Data: How To Accelerate the Re-Use of Data for Public Interest Purposes While Ensuring Data Rights and Community Flourishing*, The Governance Lab, 2020, http://dx.doi.org/10.2139/ssrn.3937638.

[23] Verhulst, Stefaan G. et al., *The Emergence of a Third Wave of Open Data: How To Accelerate the Re-Use of Data for Public Interest Purposes While Ensuring Data Rights and Community Flourishing*, The GovLab, 2020, http://dx.doi.org/10.2139/ssrn.3937638.

**Training Data:** a very large dataset used to teach a machine learning algorithm or model.[24]

**Unstructured Data:** "does not necessarily follow any format or hierarchical sequence, nor does it follow any relational rules. Unstructured data refers to masses of (usually) computerized information which do not have a data structure which is easily readable by a machine. Examples of unstructured data may include audio, video and unstructured text such as the body of an email or word processor document."[25]

---

# EXECUTIVE SUMMARY

**This paper seeks to unpack the emergent relationship between open data and generative AI. We provide a range of scenarios in which open data and generative AI could intersect and outline how open data providers and other interested parties can help make open data "ready" for those specific scenarios. By doing so, we are asking the question whether a Fourth Wave of Open Data is emerging.[26]**

Since late 2022, generative AI has taken the world by storm, with widespread use of services including ChatGPT, Gemini, Copilot, and Claude. Generative AI and large language model (LLM) applications have become a major entry point in how many individuals access, and process information. These services are being used to answer particular questions that can inform decisions as well as improve search engine functionality. As generative AI continues to transform information access, there is an opportunity to explore how generative AI can inform society in better and more ways. Towards that end, there is a need to understand how generative AI models can leverage data that is reliable, and from official sources.

Our research suggests that the intersection of open data–specifically open data from government or research institutions –and generative AI (including both LLMs and small language models) can not only improve the quality of the generative AI output but also help expand generative AI use cases and democratize open data access. For instance, when open data is used for training, fine-tuning, contextual prompt engineering, or by retrieval augmented generation (RAG) architectures, it can help improve the quality of and trust in the generative AI output (aligned with both needs and expectations) while at the same time helping to minimize hallucinations. Additionally, leveraging generative AI interfaces for open data can allow a broader group of data users to access and make decisions based on open data. However, much work needs to be done to make open data (and generative AI services) ready for these intersections from a data quality and provenance perspective.

While open data and generative AI together have great potential, understanding how exactly open data and generative AI could intersect and the requirements to make open data "ready" for those intersections remain underexplored areas. In addition, there has been limited research on whether or not generative AI can enable a Fourth Wave of Open Data–the next frontier of open data where open data is more conversational and AI ready, data quality and provenance are center stage, a whole range of new use cases from open data are feasible and there are new avenues of data collaboration. For these reasons, we have developed a "Spectrum of Scenarios"

---

framework which outlines a range of ways in which open data and generative AI can provide unique value for one another and the specifications and requirements from a data perspective to make those scenarios a reality.

## Spectrum of Scenarios

Our "Spectrum of Scenarios" framework outlines five distinct ways in which open data and generative AI can intersect in order to support open data providers and other interested parties in making open data "ready" for generative AI. Each of these scenarios includes case studies from the field and a specific set of requirements that open data providers can focus on to become ready for specific scenarios. Using this approach, we aim to make progress towards these scenarios and in the long-term become ready for all possible scenarios.

However, it is important to note that the intersections of open data and generative AI are nascent, and the landscape is rapidly evolving. These scenarios are intended to be a starting point and we intend to expand these scenarios based on the use cases available. Furthermore, whether or not open data is appropriate depends on the purpose of the use of generative AI and as such each scenario cannot be separated from the necessary problem definition phase that each LLM project should start with.

The five scenarios are summarized below:

1. **Pretraining:** Training the foundational layers of a generative AI model on vast amounts of open data
    a. Quality requirements: High volume, diverse, and representative of the desired output domain and its stakeholders, unstructured data.
    b. Metadata needs: Clear information on sourcing.

2. **Adaptation:** Fine-tuning or grounding a pre-trained model on specific open data for targeted tasks
    a. Quality requirements: High accuracy, relevance to the target task, balanced  distribution, tabular and/or unstructured data.
    b. Metadata needs: Clear labels, metadata about collection and annotation process, standardized schemas, knowledge graphs.

3. **Inference and Insight Generation:** Using a generative AI model to make inferences and extract insights from open data
    a. Quality requirements: High quality, complete, and consistent tabular data.

    b.   Metadata needs: Documented data collection methods, source information, and version control.

4. **Data Augmentation:** Leveraging open data to generate synthetic data or providing ontologies to expand training sets for specific tasks

    a.   Quality requirements: Accurate representation of real data, adherence to ethical considerations, tabular and/or unstructured data.
    b.   Metadata needs: Transparency about the generation process (including privacy compliance and ethical reviews) and potential biases.

5. **Open-Ended Exploration:** Expanding the potential of open-ended data exploration through generative AI

    a.   Quality requirements: Diverse, comprehensive, and tabular and/or unstructured data.
    b.   Metadata needs: Clear information on sourcing and copyright, understanding potential biases and limitations, entity reconciliation.

## Recommendations

As demonstrated by these scenarios, open data providers could improve the efficiency and effectiveness of generative AI, as well as expand its use for different purposes. However, much more needs to be done from a data governance perspective to make open data and generative AI ready for these use cases. In what follows, we summarize five data governance and management recommendations to help data providers and other interested parties embrace generative AI to operationalize the spectrum of scenarios.

1. **Enhance transparency and documentation:** Improving transparency and documentation of open data can not only foster ethical and responsible use throughout the data lifecycle, but can also help data holders and users to better evaluate the lineage, quality and impact of the output. This is particularly important when using open data for inference and insight generation, data and model augmentation, and open ended exploration. This may involve developing comprehensive data documentation practices such as data dictionaries,  metadata templates, and provenance standards for open data, as well as implementing standardized documentation frameworks like dataset nutrition labels and provenance tracking technologies for open data interfaces.

2. **Uphold quality and integrity:** Accelerating open data in the generative AI era demands prioritizing data quality and integrity –in particular when used for inference, data and model augmentation, and open ended exploration. There is a need for routine quality assurance processes for specific tasks (besides training), including automated or manual validation checks, and new mechanisms and tools that allow datasets to be updated based on evolving findings or changes related to the topic. Additionally, mechanisms to report and address data issues are needed to build transparency around how issues are resolved and a community around open datasets.

3. **Promote interoperability and standards:** Accelerating interoperability and implementing shared standards could help address many ongoing issues within the open data ecosystem (e.g. removing barriers to the re-use of data). Driving this shift within the ecosystem requires a coordinated approach, first adopting and promoting international data standards. This includes leading initiatives that establish data standards for data and model augmentation and synthetic data, model generated content, and other generative AI related topics and collaborating with international organizations to ensure those standards are widely recognized. Second, there is an opportunity to encourage the use of common data formats and structuring guidelines tailored to generative AI needs.

4. **Improve accessibility and useability:** Accessibility and usability are key when opening up data for secondary uses. Enhancing open data portals through intelligent search algorithms and interactive tools can help data providers and users understand what is and can be used for generative AI models. Establishing a shared space where data holders and users can come together to match the supply and demand (i.e. a data commons) can help accelerate impact in inference and insight generation and open ended exploration. In addition, clarifying and expanding sourcing information through new frameworks and guidelines can help ensure openness is appropriately balanced when handling sensitive data.

5. **Address ethical considerations:** Protecting data subjects is of the utmost priority when implementing open data for generative AI in an ethical and responsible way. Specifically, there is a need for ethical committees and comprehensive ethical guidelines around the collection, sharing and use of open data. Also, advanced privacy preserving technologies for proper de-identification and data deletion are needed as well as supporting resources to ensure these technologies are implemented responsibly.

Finally, it is essential to emphasize that this paper represents an initial foray into the rapidly evolving landscape of generative AI services, with new advancements emerging almost weekly that could address the current constraints of utilizing open data. To stay abreast of these developments and applications, we have established a living repository of examples, enabling us to swiftly update and adapt to emerging trends. Check out https://opendatapolicylab.org/ for more information.

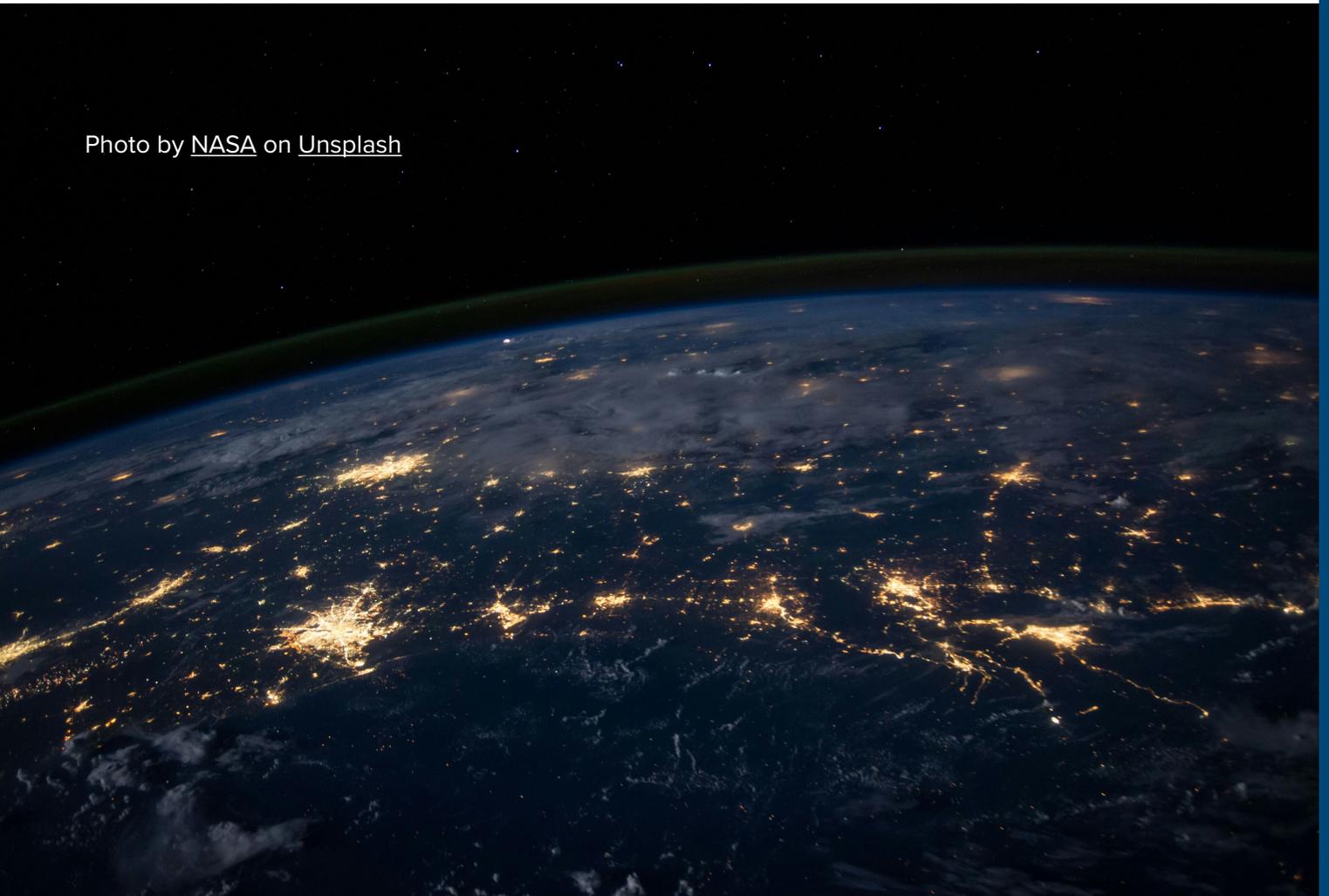

Photo by NASA on Unsplash

# A Fourth Wave of Open Data? Exploring the Spectrum of Scenarios for Open Data and Generative AI

## 1. INTRODUCTION

Since late 2022,[27] we have witnessed the rapid rise of generative artificial intelligence (generative AI) and large language model (LLM) applications, including notable services like ChatGPT, Gemini, Copilot, and Claude. These applications have quickly become staples in how many seek, access, and process information and cannot be ignored. In McKinsey's 2023 Global Survey, for instance, 33% of respondents indicated their organizations had adopted generative AI technologies into their business functions.[28] At the employee level, adoption has been staggering, with a study by Salesforce in Australia finding that 68% of Australian private sector employees across sales, service, marketing and commerce functionalities use generative AI tools to boost productivity.[29]

A similar trend emerges in the public sector as well, where a survey of 983 public servants in the United Kingdom by the Alan Turing Institute found that 22% of respondents use generative AI in their work, and 45% knew of others already using it.[30] In addition, McKinsey and Co. predicts that "generative AI could have an estimated $480 billion productivity effect on the public sector and adjacent industries."[31]

As generative AI continues to play a dominant role in how many access information, there is an opportunity to better understand how generative AI can inform society in a meaningful way. Towards this end, there is an opportunity to explore the ways in

---

which generative AI (including both LLMs and small language models) can leverage data that is reliable and from official sources to increase the quality of the output.[32]

## 1.1 The Role of Data in Generative AI

Generative AI models have long been in development, beginning with the advancement of natural language processing (NLP) models in 2010.[33] Generative AI encompasses a wide range of models that can generate content by learning concepts and patterns that are inherent in a vast array of training data, to create something new.. One example is Generative Pretrained Transformers, more commonly known as GPTs, which can enable computers to perform tasks that mimic certain human brain functions[34] such as interpreting language and generating code.

Most generative AI is powered by machine learning, which includes three main elements: algorithms, training data and models.[35] An algorithm is the process or rules that identify patterns and draw insights from training data to predict outcomes, while a model is the file that a user interacts with, which can generate insights from new data inputs.[36] In this simple example from *Scientific American*, we can see how these three elements work together: "a machine-learning algorithm could be designed to identify patterns in images, and training data could be images of dogs. The resulting machine-learning model would be a dog spotter. You would feed it an image as input and get as output whether and where in the image a set of pixels represents a dog."[37]

Data serves as the foundational backbone for generative AI models. Building and training generative AI models is a data-intensive exercise, and the specific data requirements for training, such as the quality, scale and the variety of the data for instance, often vary depending on the objectives and use cases of the model in

---

question.[38] For example, a diffusion model[39] can require between a million to a few billion image-text pairings for training.[40] In another example, ChatGPT-3.5 was trained on approximately 570 gigabytes of text data.[41]

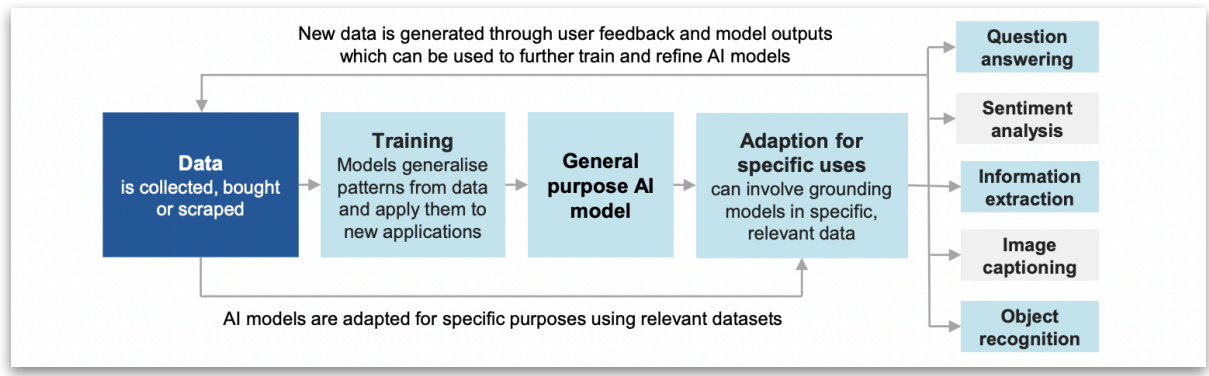

**Figure 1. The Role of Data in Generative AI**
*From the Australian Productivity Commission, this chart explains how data is both an input and an output of a generative AI system. Data is gathered and scraped from across the internet and used to train the generative AI model. From there, that model can be adapted to perform specific functions and complete different types of analysis.[42]*

## 1.2 The Potential of Open Data and Generative AI

Discourse around the potential of generative AI has been met with many concerns from the public, technology experts, activists, enterprises, and governments on the lack of transparency of generative AI models,[43] trust and mis-/dis-information (e.g. deepfakes),[44] the use of unauthorized or illegally accessed data for training (e.g.

---

38 Nvidia, "What is Generative AI?", https://www.nvidia.com/en-us/glossary/generative-ai/.

39  A diffusion model is a generative model that creates new data informed by the data it is trained on. (Source: Akruti Acharya, "An Introduction to Diffusion Models for Machine Learning", encord, August 8, 2023, https://encord.com/blog/diffusion-models/.)

40 Ibid.

41 Rita Matulionyte, "Researchers warn we could run out of data to train AI by 2026. What then?", *The Conversation*, November 27, 2023, https://theconversation.com/researchers-warn-we-could-run-out-of-data-to-train-ai-by-2026-what-then-216741.

42 Australian Government Productivity Commission, "Making the Most of the AI Opportunity - Research Paper 3: AI Raises the Stakes for Data Policy," January 2024,  https://apo.org.au/sites/default/files/resource-files/2024-02/apo-nid325426_1.pdf.

43 Saurabh Bagchi, "What is a black box? A computer scientist explains what it means when the inner workings of AIs are hidden", *The Conversation*, May 22, 2023, https://theconversation.com/what-is-a-black-box-a-computer-scientist-explains-what-it-means-when-the-inner-workings-of-ais-are-hidden-203888.

44 Ryan-Mosley, Tate. "How Generative AI Is Boosting the Spread of Disinformation and Propaganda." MIT Technology Review, October 4, 2023. https://www.technologyreview.com/2023/10/04/1080801/generative-ai-boosting-disinformation-and-propaganda-freedom-house/.

Books3),[45] the risk of hallucinations and the quality of outputs,[46] the potential to run out of training data,[47] and more (see Appendix 2).

Using open data to enhance existing efforts could not only help address these challenges, but also has the potential to democratize access to open data at the same time. The intersection of open data–in particular open government data, statistical data and open research data–and generative AI can provide value in the following ways. More information on the barriers in achieving these areas can be found in the next section of this document.[48]

A. **Elevate data user experiences through new interfaces:** Much of the open data that currently exists is published in formats that only those with technical skills can interact with. This is particularly the case for open government data. Generative AI powered interfaces can allow users to interact with complex datasets through natural language processing, visualizations, and personalized data exploration tools–thus expanding open data access to a broader group of users.

B. **Improve the quality of generative AI outputs:** The integration of high-quality, open data (specifically from governments and open research organizations) can increase the accuracy, reliability, and trust in generative AI outputs when used for inference. Nonetheless making the data sources available does not necessarily mean the data user will check them, demonstrating the need for mechanisms to support the evaluation and validation of outputs.

C. **Increase the breadth and use of generative AI:** By leveraging the diverse datasets provided by governments, official statistics, and open research initiatives, generative AI can be used to understand and generate content for a wider range of topics, use cases and sectors, e.g. from healthcare and education to climate change and economic development, as well as a wider range of world views. This broadens the applicability and relevance of generative AI to more diverse use cases for the public good.

---

D. **Foster innovation and economic growth:** Providing access to open data for generative AI can fuel innovation by providing entrepreneurs, policy makers, researchers, and developers with the raw materials needed to create new AI-driven applications and services. This can lead to the development of novel solutions to societal challenges and drive economic growth.

E. **Increase government transparency and accountability:** By integrating open government data specifically, it can become easier for government actors and the public to track and understand government actions, decisions, and policies—especially those that include technical concepts. This can lead to greater transparency, as AI can help to highlight trends, anomalies, and insights that might otherwise go unnoticed, thereby enhancing governmental accountability. However, additional work would be needed in order to motivate these actors to ask generative AI applications the right questions.

F. **Facilitate evidence-based decision-making:** Generative AI and open government data, can provide policymakers, businesses, and individuals with deeper insights and analyses about the public sphere, leading to more informed decisions. Additionally, as noted in a recent article in the Stanford Social Innovation Review, decision-makers may be able to outsource certain types of low risk decisions to an AI to help them refocus on other key issues (e.g. allowing an AI to develop meeting agendas).[49] This could help improve policy outcomes, business strategies, and even personal choices related to health, education, and finance.

G. **Increase inclusivity, multiperspectivity, and accessibility:** Generative AI can convert open data into more accessible formats like simplified texts, audio explanations, and visual aids. This promotes inclusivity and ensures that more people can benefit from valuable information as well as from the technologies built on top.[50]

H. **Enhance data literacy:** The intersection of open data and generative AI can serve as a powerful tool for education and awareness, helping to improve data literacy among the general public. By generating codebooks for analysis or presenting data in more engaging and understandable ways, people can better grasp complex issues and contribute more meaningfully to public

---

discourse.[51] However, it is important to note that data literacy will not improve automatically, intentional skill development is required.

I.   **Strengthening collaborative research:** Open government data and open research data can foster collaboration among researchers, developers, and policymakers. When combined with generative AI, it facilitates more robust research and development efforts, leading to innovations that can address societal challenges more effectively.

Open data from the public sector and research organizations can become a key resource for generative AI. It can increase the quality of the generative AI output through a broader volume of diverse datasets. However, whether or not we are moving towards a "Fourth Wave of Open Data" has yet to be determined. Among the four issues to determine the Fourth Wave include: Is open data becoming AI ready? Is open data moving towards a data commons approach? Is generative AI making open data more conversational? Will generative AI improve quality and provenance?

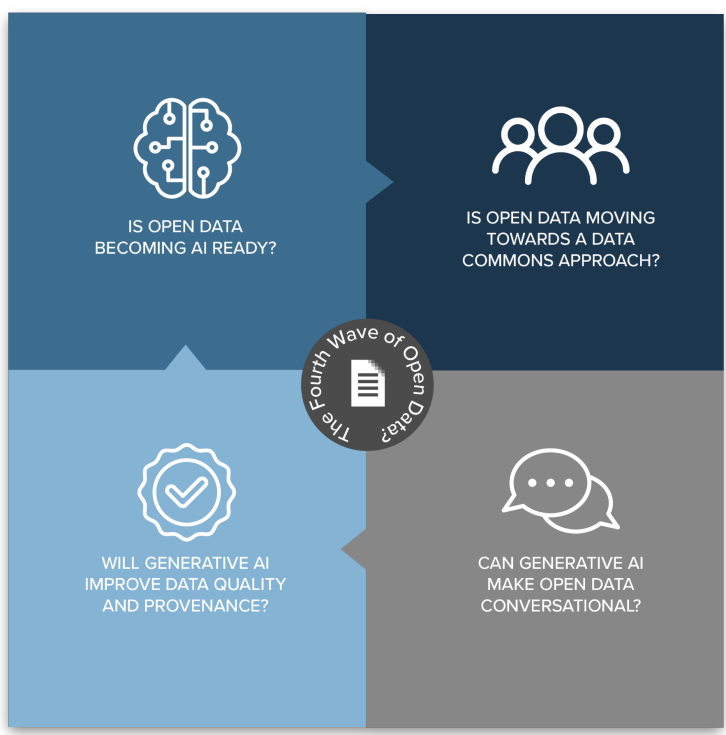

***Figure 2. The Fourth Wave of Open Data?***
*The figure above summarizes the possible components of a new Fourth Wave of Open Data.*

---

## 1.3. The Challenges of Open Data and Generative AI

The integration of open data with generative AI holds immense promise for innovation and development across various sectors. However, significant barriers impede the full realization of this potential, primarily due to the varying quality, accessibility, and compatibility of open datasets. While certain repositories like the Wikipedia knowledge hub and Google's online patent database have been instrumental in advancing generative AI, many open government and research datasets fall short of the necessary standards for effective utilization.[52] This discrepancy underscores the need for understanding the requirements across the spectrum of open data applications - as developed further below.

In the meantime, we list here the main challenges we have identified thus far. These challenges are directed both to open data providers and generative AI platforms.

A. **Quality and standardization challenges of open data for specific tasks:** The effectiveness of generative AI for tasks such as fine-tuning or inference is impacted in part by the quantity, quality and relevance of the data. Datasets that lack volume, precision, depth, or relevance can lead to suboptimal AI performance, manifesting as inaccuracies, biases, or irrelevant outputs. However, for tasks like pretraining, quality and standardization are less important and ensuring there is a sufficient volume and diversity of unstructured data is key.

B. **Interoperability and integration:** Another hurdle is the interoperability of datasets. Open data often exists in silos, each with unique formats and standards, making it challenging to integrate diverse datasets into a cohesive training corpus. Achieving interoperability requires concerted efforts to adopt universal data standards and formats that facilitate seamless data sharing and utilization across different platforms and systems.

C. **Clear information on sourcing for data and model augmentation:** Clear information on provenance and sourcing are essential to maintain transparency, trust, and accountability in the use of open data for Retrieval Augmentation Generation (RAF) architectures (or in context learning for prompt engineering) specifically. This involves establishing robust frameworks that not only track the origins of data but also ensure that contributors are duly recognized, where applicable. Such frameworks can encourage more data holders to share their resources, thereby enriching the open data ecosystem.

---

[52] Kevin Schaul et al., "Inside the secret list of websites that make AI like ChatGPT sound smart", *The Washington Post*, April 19, 2023, https://www.washingtonpost.com/technology/interactive/2023/ai-chatbot-learning/.

This requires setting up mechanisms in AI models that leverage RAG architectures to prioritize open government and open research data.

D. **Ethical, legal, and security challenges:** There are many ethical, legal, and security challenges when considering publishing open data, for use in generative AI that need to be addressed. The outputs created by generative AI must also be considered. Issues such as privacy, confidentiality and intellectual property rights should be considered. Data that is published as open data, and outputs that are generated by generative AI must respect the rights and expectations of data subjects and creators. Additionally, robust risk management plans are needed that take into account all stakeholders.

E. **The evolving nature of generative AI:** The generative AI landscape is nascent and rapidly evolving based on the latest technological advancements. There is a need to continue monitoring the generative AI landscape and ensure all intersections are appropriate and reflect current ethical, legal, and privacy challenges.

F. **Cost considerations:** Training generative AI applications can be costly. OpenAI's GPT4 cost of training is estimated around $100 million USD and models costing $10 billion could emerge in 2025.[53] Also, once the generative AI application is developed, maintaining it can have significant cost implications. As such, before generative AI applications are implemented, it is first important to understand whether the specific use case could be addressed more efficiently using other technologies or methods.

## 1.4 Our Focus

In what follows we examine a Spectrum of Scenarios where open data intersects with generative AI–specifically open government data, official statistical data, and open research data. Within each of these scenarios, we suggest a range of examples and specifications for the scenario to be operationalized in the public's interest. We then provide the requirements and differences across scenarios for them to act as a diagnostic tool. We conclude by providing a range of recommendations to help data holders and users improve access to, and gather greater insights from, open data. We hope that these scenarios serve as jumping-off points for new open data and generative AI collaborations for the public good.

---

# 2. METHODOLOGY

This paper was developed by combining The GovLab's existing work on generative AI with our Open Data Action Labs methodology—a series of design sprints with industry experts to understand the challenges and opportunities related to a topic.

We began by conducting a review of the current state of generative AI and open data. In May 2023, The Open Data Policy Lab hosted a panel discussion focused on how generative AI can be used for open data and vice versa. The panel brought to light several insights about how generative AI might be applied for open data as well as existing case studies and possible risks of its widespread application.[54] To complement this effort, The Open Data Policy Lab team conducted an in-depth literature review on the current state of generative AI and open data and aggregated key themes within a selected readings list.[55]

Through this work, we found that generative AI offers great potential for open data, but there are several ways in which open data and generative AI might intersect. Also, significant barriers remain to the implementation of these efforts. Additional work is needed to understand not only the opportunity space (i.e. what it takes to develop new interfaces), but also how to overcome key issues including interoperability, the harmonization of standards, sourcing, and attribution.

The team conducted two Open Data Action Labs or small group deliberations and ideations with a diverse group of experts from the public sector, social sector and AI ecosystem to unpack these topics further and crystalize how generative AI might play a role in a Fourth Wave of Open Data. Through two 90-minute sessions, we aimed to understand: (1) how open and statistical data can be made accessible for LLMs, and (2) how generative AI platforms can work together with statistical and government agencies.

The first Lab focused specifically on how open data can become ready for LLMs and how governments and statistical agencies can develop new interfaces for open data. Through these discussions we found that not only is open data not yet ready for generative AI, but also there is a need to specify what exactly open data is becoming ready for. Given this finding, the Open Data Policy Lab drafted a range of scenarios in which open data can provide a unique value for generative AI and used these scenarios as the basis for the second Action Lab. During the second Action Lab, the

---

expert group deliberated on the scenarios model and explored what it would take to operationalize each of them.

Lastly, the Open Data Policy Lab team conducted a series of 30-minute semi-structured interviews with domain experts to dive deeper into specific topics raised during the Action Labs. In what follows we provide a summary of the outcomes of these exercises.

Photo by Miguel Ã. Padrinan on Canva

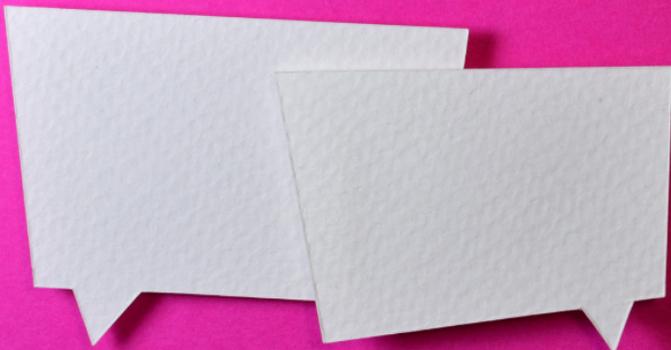

# 3. A SPECTRUM OF SCENARIOS OF OPEN DATA FOR GENERATIVE AI

In order to make open data ready for generative AI, we first need to understand "ready for what?" There are several ways in which open data and generative AI could intersect–each having its own requirements, specifications, and considerations that need to be taken into account. For example, using open data as training data for a generative AI interface would have different metadata, provenance and data quality standards than using generative AI to reason over open data and develop new insights.

"Data quality for example is very contextual. It depends on what you are going to use the data for. Some data that might be high quality in one particular domain for one particular task, may actually be low quality in another domain for another task. So we need to be very clear about what it is we are trying to accomplish."
- Harold Booth, Computer Scientist at the National Institute of Standards and Technology (NIST)

"It is important to say 'we don't have enough of a specific kind of data for certain use cases.' I would prioritize the creation, acquisition and generation of data (always with limited resources and time) so that people can focus on getting the most value out of any data we decide to invest into." - Michael Tjalve, Chief AI Architect, Tech for Social Impact at Microsoft Philanthropies

In what follows we provide a "Spectrum of Scenarios" of open data applications in generative AI. In each scenario, we describe one specific way in which open data and generative AI could intersect as well as the technical requirements for this scenario and concrete use cases. Through these scenarios, we aim to enable both providers and users to determine what open data is currently fit for what scenarios and how to upgrade it to accommodate other scenarios and goals. By doing this, we can take steps towards developing a common language for describing the scenarios and better understanding the specifics that would need to be in place to get specific types of open data ready for one scenario over another.

It is important to note that the intersections of open data and generative AI are nascent. These scenarios are intended to be a starting point and we intend to expand these scenarios based on the use cases available.

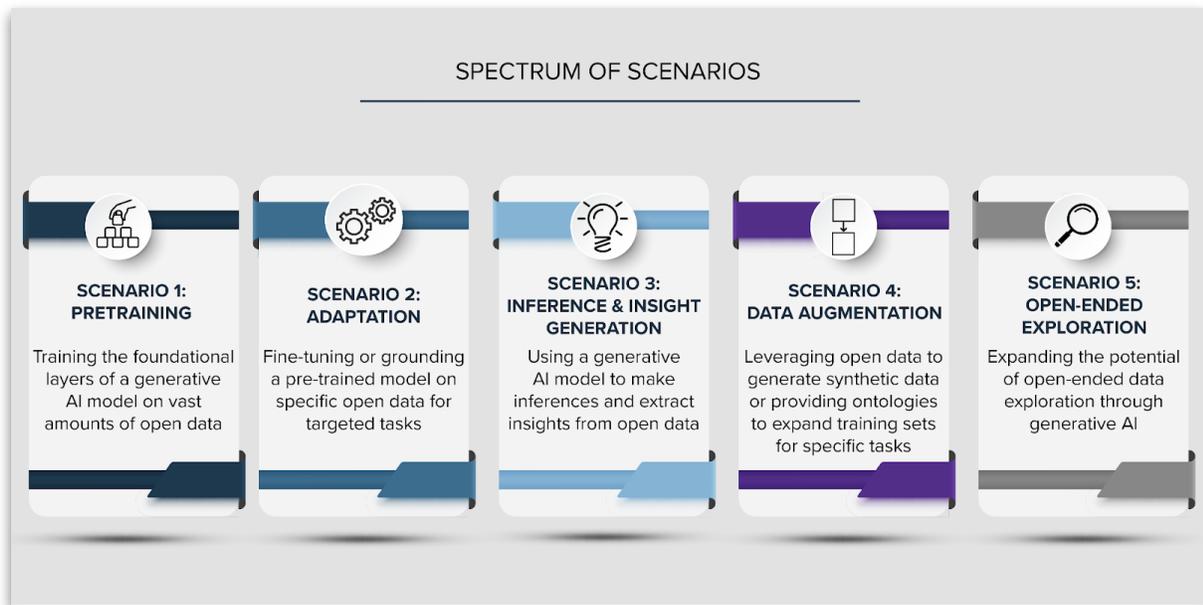

**Figure 3. Spectrum of Scenarios**
*This figure provides a visual summary of the five scenarios and primary functions of each scenario. These functions are detailed in the following sections of this report. The scenarios include: (1) Pretraining, (2) Adaptation, (3) Inference and Insight Generation), (4) Data Augmentation, and (5) Open-Ended Exploration.*

## 3.1 Scenario #1: Pretraining

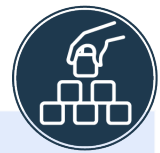

**Scenario #1 Summary**

- **Function:** Train the foundational layers of a generative AI model on vast amounts of open data.
- **Quality requirements:** High volume, diverse, and representative of the desired output domain and its stakeholders, unstructured data.
- **Metadata needs:** Clear information on sourcing.

Across sectors, we have seen many researchers, business leaders, and others raise concerns about a lack of transparency of generative AI models.[56] Generative AI technologies continue to be called into question for providing inconsistent answers to

---

questions,[57] giving incorrect information,[58] and even spreading conspiracy theories.[59] These issues start with the model, but in most cases the datasets used to train these models are scraped from across the internet and not known to the public. While there are efforts to increase transparency in generative AI documentation,[60] more can be done in order to build transparency on what types of data are included across generative AI platforms across the sector. Including open data as training data for the foundational layers of generative AI models can increase the diversity of datasets used in training and increase the transparency and accuracy of the models produced.

Towards this end, scenario one focuses on making unstructured open data ready to be leveraged by developers and data scientists to improve their model's ability to interpret prompts and generate relevant content. This would involve publishing large quantities of diverse open datasets representative of the target output domain.[61] For example, in terms of open government data, this might entail leveraging existing large national level open datasets including environmental imagery or geospatial data (as opposed to subnational level datasets) as well as unstructured elements of these datasets such as unstructured text and imagery or reporting. Thus, continuing to build upon existing efforts to open up large datasets for general purposes is integral.

For this scenario to be operationalized, open data providers would need to publish as much data as possible—emphasizing human produced unstructured data. The accuracy of this data is claimed to be less critical. There is a need for clear information on the data sources and interdisciplinary collaboration to ensure a wide range of open datasets are published both for general purposes and an array of potential use cases. Testing these training sets in real world settings is critical to ensure they are impacting the model outputs.

---

"If we look at the use of open government data [...] for pretraining, for example, that's about quantity but also about quality. Across sectors, across ministries, across departments, across borders, the purpose would be to say, let's make sure that the availability in terms of quantity is high, [...] meaning quality translates into timely release." - Barbara Ubaldi, Head of Digital Government and Data Unit and Deputy Head, Division on Open, Digital and Innovative Governments, OECD

"When you look at licenses for data, it's very all or nothing. The data may be completely open to the public. [...] but also not very findable or interoperable, making it not very re-usable." - Ashley Farley, Program Officer of Knowledge & Research Services at the Bill & Melinda Gates Foundation

### 3.1.1 Case Studies

#### NASA's HLS Project

NASA's Harmonized Landsat Sentinel-2 (HLS) project provides access to satellite data from four NASA and US Geological Survey (USGS) sensors around the globe.[62] The satellite data is complemented by auxiliary atmospheric data in real time to achieve higher quality insights.[63] The HLS dataset was used to train NASA and IBM's watsonx.ai geospatial foundation model, which can be used to rapidly develop AI systems to provide maps and analytics around natural disasters and major environmental changes for instance.[64] By providing open access to both the underlying dataset and a foundation model, researchers at NASA and IBM aim to expand access to AI technologies to help drive innovation in tackling climate and Earth science problems.[65]

#### ELMo

ELMo, or the Embeddings from Language Models, is an open source model created by a team of AI researchers at the University of Washington and the Allen Institute for Artificial Intelligence.[66] ELMo supports NLP systems by converting words into numbers, which are then used to train machine learning models.[67] Their model was

---

transformative as it was one of the first systems that was able to differentiate between word meanings based on the context of their use.[68] This is invaluable when it comes to interpreting and generating content from text-based prompts. The original ELMo model was trained on the 1 Billion Word Benchmark, which is a publicly available training dataset of nearly 1 billion words for statistical language models developed by researchers at Google, the University of Edinburgh and Cantab Research Lab.[69] Developers have found that the ELMo 5.5B model, which was trained on a more diverse set of data, including Wikipedia data and publicly available data from news platforms, is more effective.[70]

| | Source | Nearest Neighbors |
|---|---|---|
| GloVe | play | playing, game, games, played, players, plays, player, Play, football, multiplayer |
| biLM | Chico Ruiz made a spectacular play on Alusik 's grounder {. . . } | Kieffer , the only junior in the group , was commended for his ability to hit in the clutch , as well as his all-round excellent play . |
| | Olivia De Havilland signed to do a Broadway play for Garson {. . . } | {. . . } they were actors who had been handed fat roles in a successful play , and had talent enough to fill the roles competently , with nice understatement . |

***Figure 4. ELMo Training Example***
*The table above shows the nearest neighbors to the word "play" as identified by a model without contextualization (GloVe) and with contextualization (biLM).[71]*

***Additional Examples:***
- Phi-2 is an open-source small language model with 2.7 billion parameters that demonstrates outstanding reasoning and language understanding capabilities. With its compact size, Phi-2 is an ideal model for researchers to explore mechanistic interoperability, safety improvements, and fine-tuning tasks.[72]

---

## 3.2 Scenario #2: Adaptation

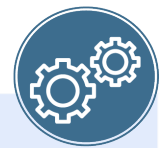

**Scenario #2 Summary**

- **Function:** Fine-tuning or grounding a pre-trained model on specific open data for targeted tasks.
- **Quality requirements:** High accuracy, relevance to the target task, balanced distribution, tabular and/or unstructured data.
- **Metadata needs:** Clear labels, metadata about collection and annotation process.

Generative AI models are typically trained on a large volume of diverse unstructured data to accomplish general tasks (more commonly known as "built for general-purpose"). While general-purpose LLMs can provide responses to different queries, these models often need to be adapted in order to complete specific tasks or to understand different domains that may not have been included in the training data.[73] For example, let's say you are seeking to understand the current state of crop disease in the state of California, a pretrained LLM adapted using agricultural data may have a higher quality output than a general-purpose LLM. What if you are looking for images of those crops? The LLM may need to be further adapted to produce images as opposed to its default output (e.g. text).[74]

Generative AI models may be adapted and complemented using fine-tuning or RAG architectures.[75] First, fine-tuning involves training one of more aspects of a pretrained model on specific open datasets.[76] Fine-tuning can be adjusted to different requirements and produce outputs quickly—thus ideal for chatbots requiring rapid user responses. Second, RAG involves setting up a retrieval system for large open datasets and embedding it in the generative AI model.[77] The idea is to steer the LLM to prioritize open data from the retrieval system over the generic dataset it was trained on.[78] RAG requires a larger volume of open data and computational capacity

---

than fine-tuning and is "more suited for tasks where contextual or factual information is crucial."[79] The data must be highly accurate, up-to-date, and clearly labeled. This includes clear metadata about the collection and annotation process. RAG is slower to produce outputs, but can provide valuable contextual information.[80]

For both approaches, the open data used must be relevant for the targeted task and representative of the problems you are seeking to solve. The data can be tabular or unstructured. Also, standardized schemas and knowledge graphs are beneficial. Understanding the labeling process is key in ensuring the quantity of each type of data is properly balanced. Finally, the two methods can be complemented by using open data to simply provide additional context when using prompt engineering.

> "The notion of [making] the data available in a machine readable form should also include, importantly, the metadata and related elements." - Robert McLellan, Principal, McLellan CIO Consulting Services
>
> "The data provenance standards—a baseline of metadata around data sets—were developed with practitioners from many corporations, large and small. As these corporations start pushing for data provenance standards to be used, the needle can start to move, and that can incentivize government potentially to do the same." - Kristina Podnar, Senior Policy Director, Data & Trust Alliance

### 3.2.1 Case Studies

#### *LLaMandement*

LLaMandement is a pretrained LLM fine-tuned by the Government of France that aims to support administrative agents in analyzing and drafting summaries of legal bills developed in the French Parliament for other ministries and departments. LLaMandement was a joint effort from the Directorate General of Public Finances, the Digital Transformation Delegation, Directorate of Tax Legislation, and others under the purview of the French Digital Republic Act—a law that enables public AI projects that are open by default.[81]

---

The team chose to fine-tune the existing pretrained LLM, LLaMA 70B, given its potential to process and analyze detailed legal documents in French and quickly adapt to new information. The fine-tuning process involved adding more parameters to the model at a low rank (otherwise known as LORA). The team used data from SIGNALE (a platform used in the French government's lawmaking) to fine-tune the model, leveraging data from several ministries including the Ministry of Ecological Transition and Territorial Cohesion, Ministry of Culture and others (see Figure 5).[82]

The team conducted a blind review process where ten experts ranked the quality of the memoranda developed by the model and compared the rankings to benchmarks from earlier models and people. Through these processes, they found that the memoranda developed by LLaMandement ranked almost at the level of memoranda developed by people. Additionally, LLaMandement was tested for "gender, ethnicity, and political ideology" related biases using English datasets given the limited availability of French datasets and demonstrated "a neutral output across various demographic and ideological dimensions."[83]

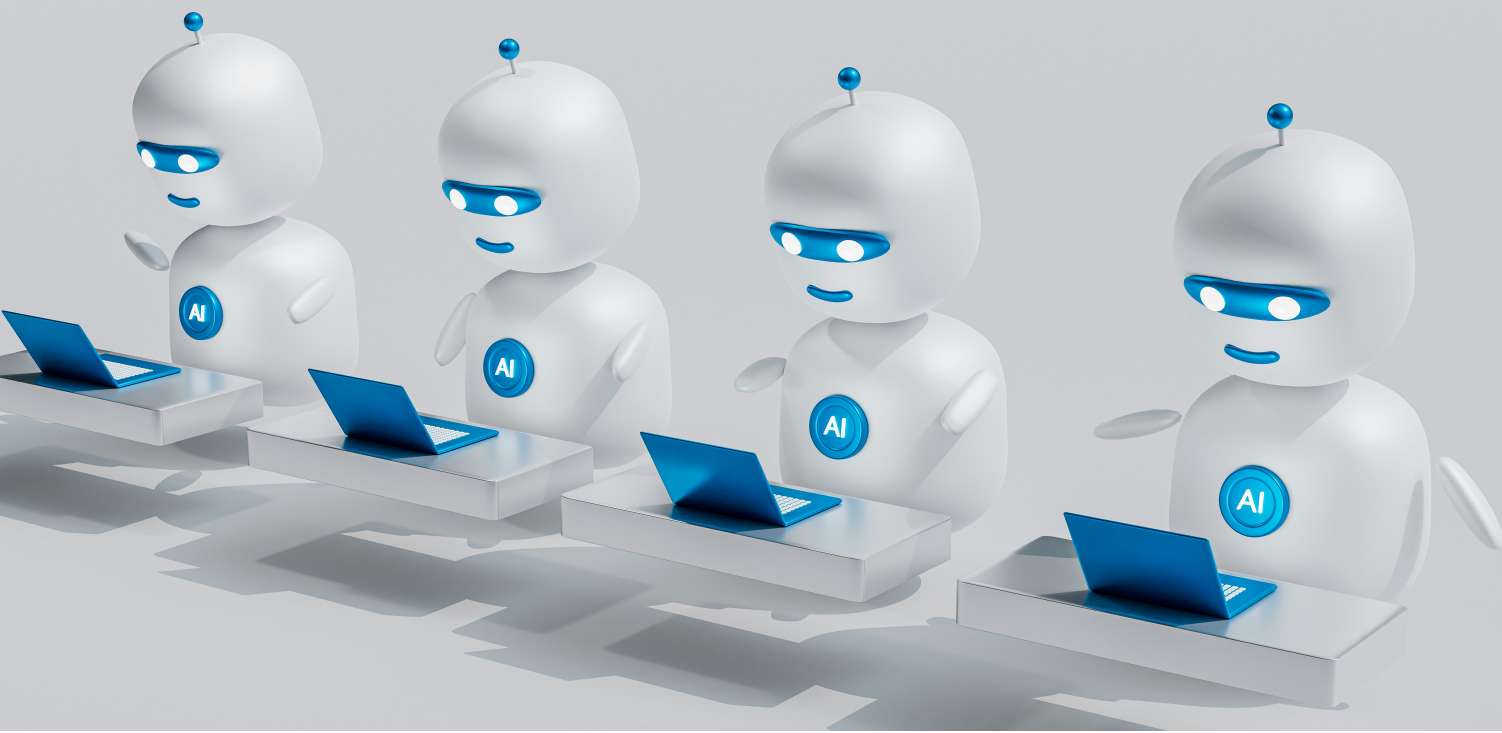

Photo by Mohamed Nohassi on Unsplash

---

[82] Ibid.

[83] Joseph Gesnouin et al., "LLaMandement: Large Language Models for Summarization of French Legislative Proposals", *arXiv*, January 29, 2024, https://doi.org/10.48550/arXiv.2401.16182.

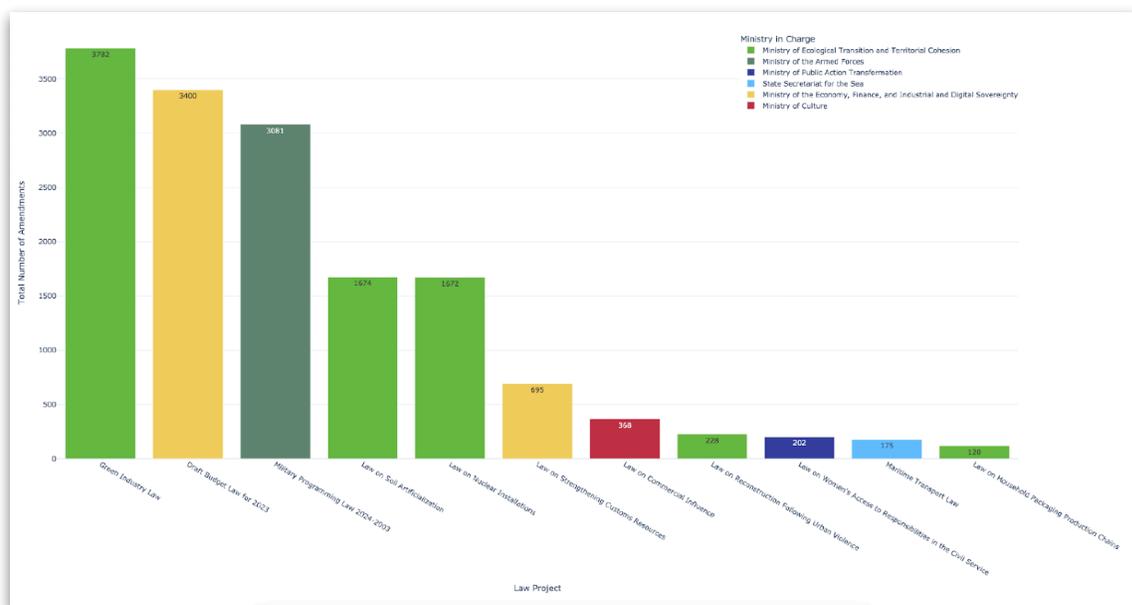

***Figure 5. LLaMandement Data Sources For Fine-Tuning***
*The figure above summarizes the breadth of the data sources used to fine-tune*
*LLaMA 70B. As demonstrated by this figure, some of the data sources were from the*
*Ministry of Ecological transition and Territorial Cohesion, Ministry of Armed Forces,*
*and others.[84]*

### Google's Med-PaLM2

A team at Google Research fine-tuned Google's general-purpose PaLM2 LLM for the medical field.[85] This fine-tuned LLM focuses on answering questions about medical exams, research, and other related topics.[86] Med-PaLM2, the most recent iteration launched in March 2023,[87] aims to generate more accurate answers to complex medical questions that require detailed answers. Med-PaLM2 is fine-tuned using "publicly available question-answering data and physician writing responses" including MedQA and MedMCQA among other datasets.[88] Med-Palm2 was evaluated using the benchmark, MultiMedQA, a group of datasets about different official exams, research, and typical patient questions. Med-PaLM2 achieved 86.5% accuracy on United States Medical Licensing Examination questions.[89]

---

## Can incontinence be cured?

**Clinician answer**

Urinary incontinence can be cured in some cases, but this depends on its cause. If it's due to an underlying urinary tract infection or constipation, treating these issues will immediately stop the incontinence. If it's due to muscle weakness, such as that caused by childbirth, strengthening exercises to build muscle tone will help and can even cure it completely, over time. If it's due to an underlying nerve problem, damage to the spinal cord, or a structural abnormality, this is usually much harder to treat and will need to be assessed on a case-by-case basis.

**Med-PaLM 2 answer**

Incontinence means the accidental loss of urine or stool. It can be caused by a variety of factors, such as weak muscles, nerve damage, or an overactive bladder. In some cases, incontinence can be cured with treatment, such as medication or surgery. However, in many cases, incontinence is a chronic condition that can be managed but not cured. Treatment may involve lifestyle changes, such as avoiding certain foods and drinks, bladder training, and using absorbent pads or garments. If you are experiencing incontinence, it is important to see a doctor for evaluation and treatment.

**Clinician review of Med-PaLM answer**

Reflects clinical and scientific consensus ✓   Low likelihood of harm ✓   Correct reasoning ✓

Correct reading comprehension ✓   Correct knowledge recall ✓   Relevant content ✓

No missing important information ✓   No demographic biases ✓

✓ Complete answer   ◑ Incomplete answer

***Figure 6. Med-PaLM2 Example Question***
*This image demonstrated how Med-PaLM2 answered a medical question compared to a medical professional. The figure shows that Med-PaLM2 was able to provide a complete answer to the question, "can urinary incontinence be cured?" across several dimensions.[90]*

### Additional Examples:

- OpenAssistant Conversations generates crowdsourced open datasets and fine-tuned models to accelerate research on LLMs.[91]

- Researchers from the Mayo Clinic and the University of Illinois fine-tuned LLaMA using health records to increase the efficiency of assigning patients to different diagnosis related groups for administrative purposes.[92]

---

[90] Ibid.

[91] Andreas Köpf et al., "OpenAssistant Conversations -- Democratizing Large Language Model Alignment", *arXiv*, October 31, 2023, https://doi.org/10.48550/arXiv.2304.07327.

[92] Hanyin Wang et al., "DRG-LLaMA : tuning LLaMA model to predict diagnosis-related group for hospitalized patients", *npj Digital Medicine* 7 no. 1 (2022), https://pubmed.ncbi.nlm.nih.gov/38253711/.

- Researchers at MIT in collaboration with other institutions have developed the Data Provenance Explorer which is an online dashboard of datasets that have been used to train or fine-tune AI models. The dashboard includes information about these datasets' licenses and metadata.[93]

## 3.3 Scenario Inference and Insight Generation

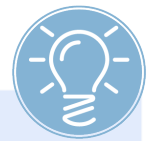

**Scenario #3 Summary**

- **Function:** Extract insights and patterns from open data using a trained generative AI model.
- **Quality requirements:** High quality, complete, and consistent tabular data.
- **Metadata needs:** Documented data collection methods, source information, and version control.

Once an LLM is trained to analyze data and extract insights, it can be used to power generative AI interfaces for open data. These interfaces have the ability to process language-based prompts and questions related specifically to the open dataset, and to generate appropriate insights and responses. They can help reason over data and improve the functionality of existing open data search engines. Given their defined scope (focused on certain datasets), these interfaces are usually more accurate and less prone to hallucinations and biases when compared to more broadly applicable models.[94] An example of this scenario in action would be Census GPT, which runs on census data from the 2021 American Community Survey to answer questions related to crime, demographics, education, income and population trends in the United States.[95] However, this example is at the early stages and the functionality is in need of improvement. Other potential use cases around generative AI and insight generation would be models that are able to analyze crime patterns, predict disease outbreaks, or identify emerging social media trends based on tabular open data.

---

[93] MIT Media Lab. "Data Provenance for AI." Accessed March 26, 2024. https://www.media.mit.edu/projects/data-provenance-for-ai/overview/.

[94] Amra Dorjbayar, Interview re: open data and generative AI. Zoom, February 14, 2023.

[95] "Census GPT", https://censusgpt.com/.

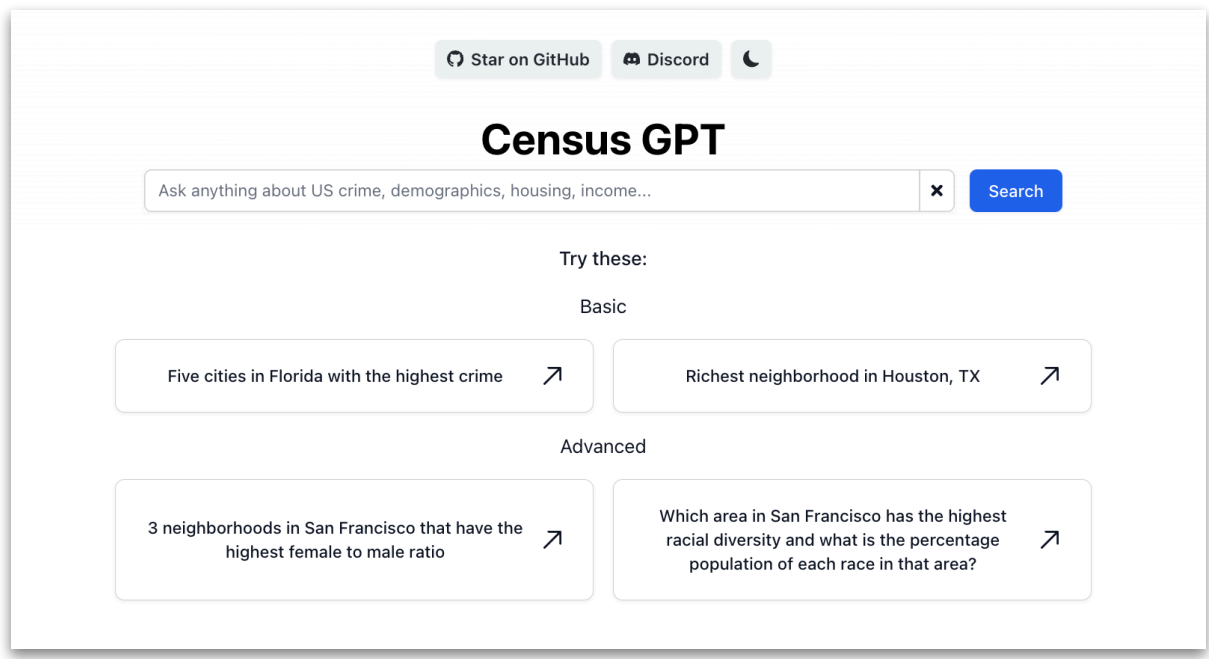

**Figure 7. Census GPT Interface**
*This screen capture shows Census GPT's interface along with sample prompts the user can ask the model, including cities in Florida with the highest crime rates, richest neighborhoods in Houston, and neighborhoods in San Francisco with the highest female to male ratio. In addition to these prompts, users can type in other questions that the interface can address.[96]*

When it comes to open data for these models, there are a number of factors to consider. First, it is important that the data is complete (including all required elements for the intended purpose), consistent and reliable to ensure that the model is producing accurate insights. As computer and data scientists say, 'garbage in, garbage out'. This means that underlying issues in the data, such as biases, incompleteness and out of date information to name a few, lead to low quality outputs regardless of the strength of the model. In addition to quality, data provenance and robust metadata are critical when it comes to inference and insight generation. Without a clear understanding of where the data is coming from and its surrounding context, insights gleaned from the data may be misinterpreted or misused. As such it is important that users understand how and where the data was collected, by whom and for what purpose. To that end, this scenario may also require individual consent from the data user. Metadata can help match users with the right datasets to achieve more efficient and effective outcomes.

---

### 3.3.1 Case Studies

***Wobby***

Wobby is a generative AI-powered interface that accepts language-based queries and prompts related to a specific open dataset to produce responses in the form of summaries and visualizations.[97] The platform is focused primarily on democratizing access to open government data, and currently hosts datasets from organizations like Statbel (Belgium's national statistical office), Statistics Netherlands and Eurostat, as well as data from intergovernmental organizations like the World Bank.[98] By combining data analytics with coding knowledge, Wobby aims to expand access to open government data among a broad audience.[99]

> "It's really hard, if not impossible to just, grab random files on Kaggle or [...] [other open source platforms]. How old is the data? I don't know. Is it from a reliable source? I don't know. So we're starting from the reliable ones with an API that we can discover in batch." - Amra Dorjbayar, Co-Founder & CEO of Wobby

When sourcing the data available on Wobby, the team prioritized a few conditions. They sought datasets with strong APIs that operated on a single system. This made it easier to collect and share the data on their platform. They also partnered with open data publishers, limiting their data sources solely to governments and intergovernmental organizations to avoid data provenance and quality challenges.[100]

| maand | mannen | vrouwen | year | |
|---|---|---|---|---|
| ⏱ Temporal | # Numerical | # Numerical | ⏱ Temporal | |
| Month and year in the format yyyy-mm. | Number of men in the given month and year. | Number of women in the given month and year. | Year corresponding to the data in the format yyyy. | |
| 2024-02 | 21 | 1 | 2024 | |
| 2024-01 | 66 | 12 | 2024 | |
| 2023-12 | 23 | 6 | 2023 | |
| 2023-11 | 58 | 7 | 2023 | |
| 2023-10 | 38 | 9 | 2023 | |
| 2023-09 | 64 | 24 | 2023 | |
| 2023-08 | 48 | 14 | 2023 | |
| 2023-07 | 28 | 6 | 2023 | |

Monthly Gender-wise Employee Recruitment in Belgium
74 records
This dataset provides monthly data on the number of new hires by gender in Belgium from 2018 to 2023. It allows us to see how the number of male and female hires has changed over time. This data can be used to identify trends in hiring practices and to monitor progress towards gender equality in the workplace.

---

[97] "Wobby", https://wobby.ai/.

[98] Ibid.

[99] Amra Dorjbayar, Interview re: open data and generative AI. Zoom, February 14, 2023.

[100] Ibid.

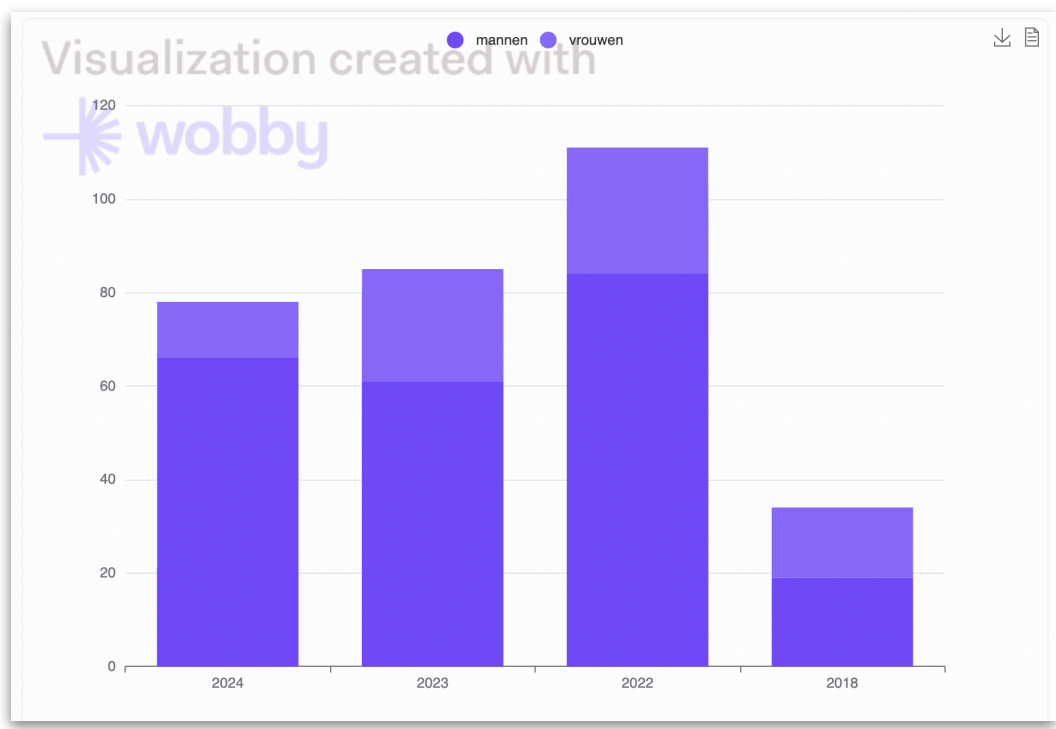

***Figure 8. Wobby Bar Chart***
*The images above generated by Wobby show monthly gender-wise employee recruitment in Infrabel (a Belgian train infrastructure company) between 2018 and 2024. It shows the data in both a table and bar chart format. [101]*

### The IMF's StatGPT

To help improve the accessibility and usability of their open data platform, the International Monetary Fund (IMF) is prototyping a new generative AI tool that they are calling StatGPT.[102] StatGPT will serve as a user interface that can process natural language requests and commands into a Statistical Data and Metadata eXchange (SDMX) data query to identify and source relevant datasets from the IMF's data repository.[103] In addition to collecting datasets upon request, StatGPT will help users find best-fit indicators, visualize data as tables and charts, and generate Python code for their analysis.[104] Since the tool operates on an in-house dataset, it is also able to enforce data access rights and accurately represent data provenance.[105]

---

[101] This image was generated using Wobby's platform. Learn more at: https://wobby.ai/.

[102] UNECE High Level Group on Modernisation of Official Statistics (HLG-MOS) Modernisation Groups, "Large Language Models for Official Statistics: HLG-MOS White Paper", UNECE, December 2023, https://unece.org/sites/default/files/2023-12/HLGMOS%20LLM%20Paper_Preprint_1.pdf.

[103] Ibid.

[104] Ibid.

[105] Ibid.

***Additional Examples:***

- In Switzerland, a collaboration between the Federal Statistical Office's Data Science Competence Center, numerous regional statistical offices, the Zurich University for Applied Sciences and the Digital Public Services Switzerland is working to develop StatBot.swiss—a chatbot that will be able to answer questions related to open administrative data.[106]

- Data Commons is in the process of building a natural language interface, powered by a series of LLMs, to improve the accessibility and usability of their vast collection of data. The new system will be able to create data visualizations in response to text-based prompts, as well as provide information about the data sources used to inform its analysis.[107]

## 3.4 Scenario #4: Data Augmentation

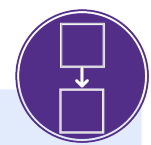

**Scenario #4 Summary**

- **Function:** Leveraging open data to generate synthetic data or providing ontologies to expand training sets for specific tasks.
- **Quality requirements:** Accurate representation of real data, adherence to ethical considerations, tabular and/or unstructured data.
- **Metadata needs:** Transparency about the generation process and potential biases.

Data augmentation helps expand training sets to include more data representative of different topics and improve the quality of the generative AI output.[108] There are several approaches that may be utilized for data augmentation. Below we provide a summary of how open data may be used in two approaches: synthetic data generation and ontologies.

First, synthetic data is increasingly used in AI models as training data.[109] Synthetic data refers to "artificially generated data that mimics real data characteristics without containing any actual personal information" (see Glossary of Terms for detailed

---

definition).[110] Synthetic data can be created through several processes depending on the data type and source. For example, synthetic data may be generated by developing a model that mirrors real world data (e.g. by using a LLM), or by conducting a series of simulations.[111] Synthetic data is often cheaper to produce than real world data and can be used to protect the privacy of the data subjects or minimize bias in datasets. However, synthetic data can still result in harm to the data subjects if it is used as a way to avoid privacy compliance processes and regulations.

Second, open government and open research data providers may supply ontologies as open data as part of data augmentation processes. Ontologies refer to "a description of data structure—of classes, properties, and relationships in a domain of knowledge."[112] Ontologies aim to provide a framework for describing a specific subject area.[113] Ontologies are typically used to develop or improve knowledge graphs that are applied during pretraining, fine-tuning, RAG, and other instances. In turn, they can help increase the explainability of AI systems.[114]

In this scenario, the priority is making the augmented tabular or unstructured datasets fully representative of real world data. These datasets need to be relevant, timely, and accurate for the context at hand. Also, augmented datasets must adhere to ethical considerations and all necessary privacy compliance processes, maintain transparency around the generation process, and acknowledge potential biases. Failing to disclose these aspects could lead to privacy related issues and less accurate outputs which ultimately hinders the LLMs performance in the long-term.

### 3.4.1 Case Studies

***Synthetic Australian Healthcare Data Using Synthea***

In January 2024, researchers at the Australian e-Health Research Centre within the Commonwealth Scientific and Industrial Research Organisation (CSIRO) and Macquarie University published a report on how synthetic data can accelerate access to healthcare data. In Australia, privacy concerns, high cost, and other factors continue to limit access to healthcare data. At the same time, much of the synthetic data that already exists tends to focus on North America and Europe and is not tailored to the Australian healthcare sector.[115] To this end, the team chose to leverage Synthea, an online synthetic data tool for the healthcare sector based on census data from the United States,[116] to create synthetic Australian health records.

---

[110] Willem Koenders, "Navigating the data management landscape in the age of Gen AI", *Medium* (blog), January 29, 2024, https://medium.com/zs-associates/navigating-the-data-management-landscape-in-the-age-of-gen-ai-82a5337a8c00.

[111] Khaled El Emam et al., "Chapter 1. Introducing Synthetic Data Generation" in *Practical Synthetic Data Generation*, O'Reilly Media, Inc. (2020), https://www.oreilly.com/library/view/practical-synthetic-data/9781492072737/ch01.html.

[112] Oxford Semantic Technologies. "What Is an Ontology?" Accessed April 2, 2024. https://www.oxfordsemantic.tech/faqs/what-is-an-ontology.

[113] Confalonieri, Roberto, and Giancarlo Guizzardi. "On the Multiple Roles of Ontologies in Explainable AI." arXiv, November 8, 2023. http://arxiv.org/abs/2311.04778.

[114] Ibid.

The team adapted the Synthea model to include Australian demographic data (including socio-economic factors) and hospital data. Using this model the team was able to generate approximately 117,000 synthetic health records specific to Queensland, Australia. From there, the team conducted a series of modeling exercises to understand patterns in different types of diseases. The team noted that the use of Synthea provided value in generating access to sensitive data, but there is a need for additional real world testing to ensure it accurately captures the local context.[117]

| Data set | Description |
|---|---|
| patients.csv | Demographics information of patients |
| encounters.csv | Encounter with a medical practitioner |
| conditions.csv | Conditions diagnosed at encounter date |
| allergies.csv | Presence of allergies and allergy category |
| imaging_studies.csv | Imaging studies such as x-ray, ultrasound… |
| observations.csv | Include vital signs, laboratory tests, survey… |
| medications.csv | Past and current medication with reason for prescription |
| providers.csv | Includes address and contact details of providers such as GP practices |
| careplans.csv | Reason for care plan, start and stop dates |
| procedures.csv | Assessment, screening, chemotherapy |
| devices.csv | Devices used for treatment for example (home nebuliser) |
| immunisation.csv | Immunisations history |
| organisations.csv | visited institutions such as hospitals, with address, phone… |
| payers.csv | Public and private health insurance organisations |

**Figure 9. Synthea's Synthetic Data**
*This figure summarizes the variety of datasets that the Synthea model can provide. This includes data about patient demographics to other elements in medical records.[118]*

---

***Statistics Canada's Use of Synthetic Data***

Statistics Canada conducted a pilot program around generating synthetic data for training purposes. The synthetic datasets were developed using census data containing sensitive information and provided to participants during two Hackathons with the caveat that the datasets could not be made publicly available. The organizing team noted that the synthetic datasets developed were able to maintain "the analytic utility of the original data while effectively reducing the risk of disclosure." Additionally, these datasets were able to be used by Hackathon participants for training.[119]

***Additional Examples:***

- The Government of Switzerland's I14Y metadata catalog provides a series of ontologies as open data across several domains.[120]

- The Urban Institute in collaboration with the Department of Human Services in Allegheny County, Pennsylvania and the Western Pennsylvania Regional Data Center developed a synthetic version of its dataset on service usage (to minimize privacy concerns around sensitive data) for researchers, organizations, and the public to use.[121]

- GretelAI is an online platform that supports developers in generating synthetic data. Previous clients include the Government of South Australia and the United States Department of Justice.[122]

- Researchers at the National Transportation Research Center in the United States used data augmentation to incorporate more granular elements into a truck travel dataset from the Next Generation National Household Travel Survey program for AI training purposes.[123]

- In the United States, the Office for National Statistics' Data Science Campus created a synthetic dataset using the U.S. Census Bureau's income data with the goal of testing the 2021 Census model.[124]

## 3.5 Scenario #5: Open-Ended Exploration

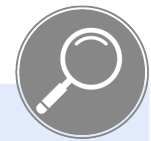

**Scenario #5 Summary**

- **Function:** Explore and discover new knowledge and patterns within open data using generative models.
- **Quality requirements:** Diverse, comprehensive, and tabular and/or unstructured data.
- **Metadata needs:** Clear information on sourcing and copyright, understanding potential biases and limitations, entity reconciliation.

The generative AI ecosystem is fast changing, presenting exciting and innovative use cases for the public good. One of the directions the field is moving towards is open-ended exploration, wherein generative AI models are used to discover new patterns and knowledge in open data.[125] For example, AI platform, Jaxon, collaborated with the US National Oceanic and Atmospheric Administration (NOAA) to develop a map of seafloor habitats in the US Caribbean by automating a previously manual and time consuming data annotation process.[126] Their new system improved the quality of data for the map and reduced their processing times.[127] Other examples of open-ended exploration could include using generative AI to create alternative text formats, visualizations, or even musical pieces based on specific datasets.

In order to achieve open-ended exploration use cases, generative AI models require significant amounts of diverse and comprehensive open data. The open data can be tabular or unstructured. A varied and comprehensive dataset is essential for exploring new hypotheses and generating unique outputs. This can also help mitigate the potential risk of bias in the data and in the model's output. Since these models aim to unearth new trends and insights from the data, it is important to have clear information on sourcing and copyright. Additionally, entity reconciliation[128] across data types and sources is needed to help accelerate interoperability efforts. These details play a key role in contextualizing insights and evaluating potential biases or limitations of the analysis. It is also important as a way to foster greater transparency and replicability in the open data and generative AI ecosystem.

---

[125] Johanna Walker et al., "Prompting Datasets: Data Discovery with Conversational Agents", *arXiv*, December 15, 2023, https://doi.org/10.48550/arXiv.2312.09947.

[126] "Mapping Seafloor Habitats with AI", Jaxon.ai, https://jaxon.ai/mapping-seafloor-habitats-with-ai/.

[127] Ibid.

[128] Entity reconciliation describes the process of "identifying entities from the digital world that refer to the same real-world entity."
Enríquez, J. G., F. J. Domínguez-Mayo, M. J. Escalona, M. Ross, and G. Staples. "Entity Reconciliation in Big Data Sources: A Systematic Mapping Study." Expert Systems with Applications 80 (September 1, 2017): 14–27. https://doi.org/10.1016/j.eswa.2017.03.010.

"I think that we should anticipate that [user expectations] will change. And there will be an expectation for natural language, NLP interfaces to accomplish all sorts of tasks. And that will become more ubiquitous. And as that occurs, there's going to be an expectation that if users can leverage these technologies that are increasingly ubiquitous to analyze public data and get meaningful results." - Oliver Wise, Chief Data Officer at the United States Department of Commerce

"When it comes to generative AI, our focus is precisely on how we can bridge the ease-of-use gap in our current methods of open data dissemination."
- Scott Beliveau, Principal Innovation Architect at the United States Patent and Trademark Office

### 3.5.1 Case Studies

***NEPAccess***

NEPAccess is a project by a team of researchers at the University of Arizona, who seek to operationalize the 1969 National Environmental Policy Act (NEPA). As part of NEPA's approach to environmental governance, the act seeks to improve citizen engagement in government decision making and use research to improve policy outcomes related to social and environmental well being.[129] Towards that end, NEPA collects and publishes data focused on environmental conditions across the country and from research done as part of federally funded projects.[130]

NEPAccess works to unlock access to this data using a generative AI model. As noted by the team, the model can support users in searching for NEPA data and reviews that were otherwise challenging to find.[131] The model does not yet have analytical capabilities and mainly focuses on retrieving unstructured NEPA data using text-based search prompts.[132] The next iteration of NEPAccess will include expanding the platform with features to help users write environmental impact assessments, new data analysis offerings, and others.[133]

***Parla***

Parla is a new AI interface that is currently being prototyped by the CityLab Berlin, as a way to improve access to the city's public administration data for both government

---

actors and the public more broadly.[134] Working as both a retrieval system and an analytical tool, Parla accesses over 10,000 public domain documents spread across different city departments, systems and formats to respond to simple natural language prompts.[135] Unfortunately without well-structured and machine-readable data, as well as high quality metadata and other factors, the system has been prone to hallucinations and other challenges related to the quality of its outputs.[136] One approach Parla is taking to mitigate this risk is ensuring that the responses provide references to sources to help improve transparency and traceability.[137]

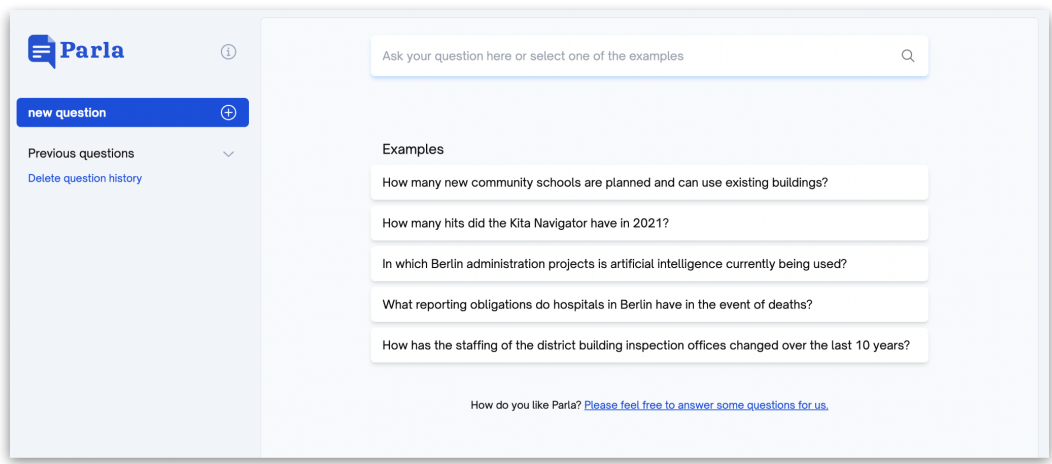

**Figure 10. Parla by CityLab Berlin**
*This figure shows Parla's interface with suggested questions for users to ask the model. On the interface, there is also the ability to view previous and new questions.[138]*

**Additional Examples:**
- The LLM on FHIR project uses LLMs to enable patients to access and interact with their healthcare data through a conversational interface with the goal of improving health literacy and reducing barriers to information.[139]

- Clay, a project by non profit organization, Radiant Earth, aims to bridge gaps that persist in environmental and Earth data through their open-source AI model, which collects and organizes diverse data sources (including open

---

government data) into a geospatial dataset that can be analyzed to achieve greater insights than what is currently possible.[140]

- Talk to the City, created by the AI Objectives Institute, works to improve collective deliberation and decision making by leveraging an LLM interface to analyze large amounts of qualitative data from responses to opinion surveys. The platform clusters similar responses and provides users with a visual and written summary of responses for rapid analysis.[141]

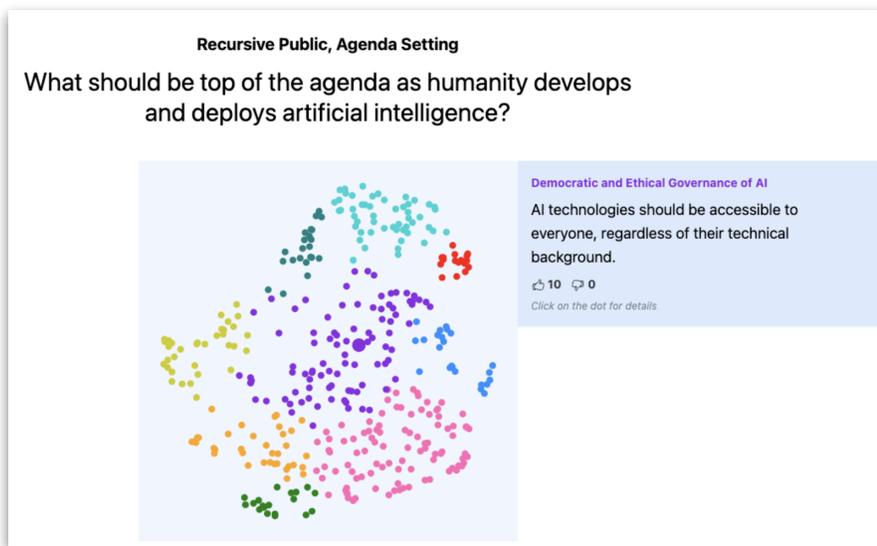

***Figure 11. Talk to the City Sample Output***
*An example of Talk to the City's clustering of citizen responses as part of a public agenda setting exercise. The figure includes the question, a scatterplot, and sample response.[142]*

---

# 4. OPEN DATA REQUIREMENTS AND DIAGNOSTIC

The "Spectrum of Scenarios" above outlines the technical requirements and concrete use cases for each scenario with the goal of enabling open data providers and users to evaluate what open data is currently fit for use for which scenarios. This comparison can also serve as a roadmap to move between scenarios towards different uses and goals. Drawing on this framework, we have identified the following requirements and differences across the scenarios to act as an initial diagnostic tool:

A. **Transparency and Documentation:** Advancements in Generative AI have expanded training capabilities to perform on a large volume of unstructured datasets, rather than always requiring well-structured and documented, tabular data. For example, Inference and Insight Generation (Scenario #3) may rely on tabular data, while pretraining (Scenario #1), data and model augmentation (scenario #4) and open-ended exploration (scenario #5) may be able to operate solely on unstructured datasets. While having structured and labeled data are crucial for many data users, applying provenance standards in generative AI training could limit development significantly.

B. **Quality and Integrity:** Data holders ought to maintain high standards for accuracy, completeness, and consistency in their data. It is also important that they implement robust mechanisms for validation for end users and error correction. RAG architectures (Scenario #2), for instance, demands higher data precision and relevance when compared to pretraining scenarios (Scenario #1), where data volume and diversity are prioritized.

C. **Interoperability and Standards:** To reduce barriers for use, data holders and users ought to adhere to established data formats and standards. This ensures broader compatibility across systems and increases the interoperability of the data. Data augmentation (Scenario #4) might necessitate more flexible or innovative data formats to accommodate the wide variety of synthetic data being developed. In contrast, traditional formats are suitable for pretraining (Scenario #1) or adaptation (Scenario #2).

D. **Accessibility and Usability:** Data should be made easily accessible and machine-readable to improve its usability for generative AI. It should also be accompanied by clear sourcing information to maintain data provenance and integrity. For example, open-ended exploration (Scenario #5) requires not only accessible data but also extensive datasets that cover a wide range of domains to enable novel discoveries, which is less critical in more focused scenarios like adaptation (Scenario #2).

**E. Ethical Considerations and Privacy:** To protect data subjects and ensure the ethical use of open data, especially open government data and statistical information, which often contains sensitive data, ethical data collection and usage practices are key. These can include practices like anonymization and synthetic data generation where appropriate. Scenarios involving sensitive information, such as healthcare data in data augmentation (Scenario #4) for example, require stricter privacy controls and ethical considerations compared to scenarios dealing with less sensitive information, like public domain text for pretraining (Scenario #1).

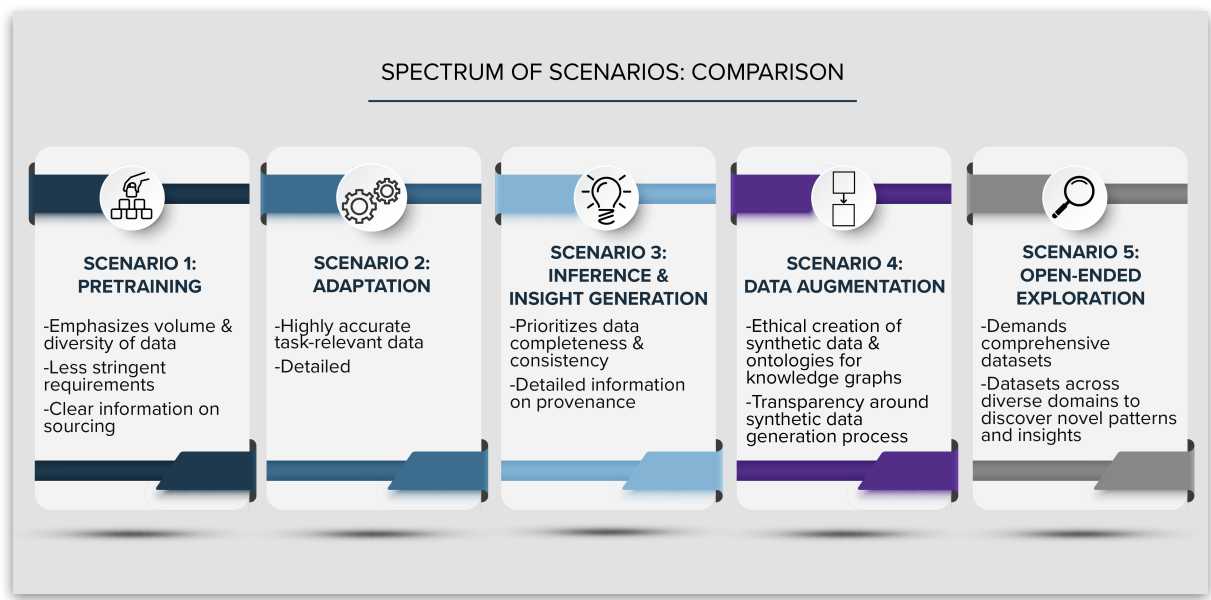

**Figure 12. Comparison across Scenarios**
*This chart summarizes the data requirements across the Spectrum of Scenarios. These requirements were outlined in the previous sections of this document.*

# 5. RECOMMENDATIONS FOR ADVANCING OPEN DATA IN GENERATIVE AI

While developing the "Spectrum of Scenarios", we learned a lot about the different requirements behind open data for generative AI. However, it is also important to think about how to advance open data in the context of generative AI. And so, drawing on lessons learned, we have developed the following recommendations for data governance and management to help data holders and other interested parties improve access to, gain greater insights from, open data.

## Enhance Transparency and Documentation

Enhancing transparency and documentation of open data can not only help ensure it is used in an ethical and responsible manner throughout the data lifecycle, but can also help data holders and users to better evaluate the value and impact of the data in question. This is particularly useful when using generative AI for open data inference and insight generation, data augmentation, and open ended exploration. Actions to promote greater transparency and documentation efforts can be divided into two scopes of action:

**A. Developing Comprehensive Data Documentation Practices:**
   a. Include advanced data dictionaries, detailed provenance information, and version histories to track changes over time on generative AI enabled open data interfaces.

   b. Explore the creation of new metadata templates tailored to the specific needs of generative AI, incorporating elements such as data lineage, ethical considerations, and intended use cases.

   c. Creating Data Provenance Standards for Open Data: There is a need for developing and establishing new standards and templates for open data focused on data provenance. These standards would make it easier for open data providers to document and share provenance information in a consistent and comprehensive manner, enhancing the overall transparency and trustworthiness of open datasets. This would be beneficial for evaluation of output - when used for RAG architectures (Scenario #2) or Inference (Scenario #3), but would likely not be feasible for training purposes (Scenario #1).

**B. Implementing Standardized Documentation Frameworks:**
   a. Adopt and adapt frameworks like Datasheets for Datasets and Dataset Nutrition Labels for open data, ensuring they include sections relevant to generative AI, such as model compatibility, potential biases, and suggested applications. Similar to the above, this would focus on adaptation and inference as opposed to training.

b. Provenance Tracking Technologies: Investing in technologies and methodologies that enable robust tracking and recording of data provenance is an area for further development. Blockchain, for example, has been explored as a means to ensure tamper-proof provenance tracking. Such technologies could provide immutable records of data origin, lineage, and usage, which is valuable in contexts where data integrity and authenticity are paramount.

c. Encourage community-driven development and adoption of these frameworks to reflect the evolving nature of generative AI technologies.

## Uphold Quality and Integrity

Upholding data quality and integrity are key when thinking about advancing generative AI for open data inference and insight generation, data augmentation, and open ended exploration. . This may involve activities such as:

**A. Establishing Routine Quality Assurance Processes:**
   a. Integrate automated and manual validation checks that are tailored to specific tasks (besides training) such as RAG or inference. .

   b. Develop dynamic updating mechanisms that allow datasets within generative AI interfaces to evolve in response to new findings, user feedback, and changes in the relevant domain.

**B. Creating Mechanisms for Reporting and Addressing Data Issues:**
   a. Implement user-friendly platforms for reporting data errors or inconsistencies, ensuring transparent communication about how these issues are resolved within open data interfaces and data augmentation processes.

   b. Deliver specific mechanisms aimed to prevent the reidentification of data subjects and users when privacy preserving processes are used on open data and uphold the right to be forgotten.

   c. Foster a community around open data sets, encouraging users to contribute to the improvement and refinement of data quality.

## Promote Interoperability and Standards

Improving the interoperability of data and promoting the adoption of shared data and metadata standards would address many long standing pain points in the open data ecosystem that prevent the efficient and effective use of open data. To drive this shift in the ecosystem, data holders could take the following actions:

**A. Adopting and Promoting International Data Standards and Schemas:**
   a. Lead initiatives to develop and establish new data standards that cater specifically to the needs of generative AI, such as standards for metadata, synthetic data, or model-generated content. The development of these standards should include processes that can engage civil society and ensure there is an ample social license to operate.
   b. Work with international bodies to ensure these standards are recognized and adopted globally, facilitating cross-border data sharing and collaboration.

**B. Encouraging the Use of Common Data Formats and Structuring Guidelines:**
   a. Develop guidelines and best practices for data structuring that consider the specific requirements of generative AI models, such as format compatibility and data richness.
   b. Promote the adoption of these practices through workshops, webinars, and educational resources aimed at data providers.

**C. Promoting the Use of Tabular Data in an Efficient and Ethical Manner:[143]**
   a. Design new formats of machine-readable tabular data along with detailed metadata to help generative AI interfaces for open data and data augmentation processes achieve greater insights.
   b. Introduce new processing methods and prompt formats to turn NLP requests into clear tasks for the LLM to run on tabular data.

## Improve Accessibility and Usability

Along with data interoperability, accessibility and usability are important factors to consider when opening up data for broader use. Generative AI can play a role in this process, however other key activities include:

---

A. **Enhancing Open Data Portals:**
   a. Invest in advanced features for open data portals, such as generative AI enabled search algorithms that understand the context and intended use of data queries.
   b. Integrate interactive tools that allow users to preview data compatibility with common generative AI models directly within the portal.

B. **Creating (and Re-imagining) Data Commons:**
   a. Establish common data spaces where data holders and users can come together to help better match supply with demand to develop new insights and achieve greater outcomes for the public good.
   b. Modernize governance models of data commons ensuring ethical use and management of data resources through new steward roles and community-informed processes.
   c. Develop equitable access mechanisms that balance openness and misuse prevention, foster active community participation, and establish quality and sustainability standards to maintain the health and usefulness of data commons in the face of evolving technological landscapes like.

C. **Clarifying and Expanding Sourcing Frameworks:**
   a. Develop new sourcing frameworks that address the unique aspects of generative AI, such as the use of data in model training and testing versus direct consumption.
   b. Provide clear guidelines and examples to help data providers select the most appropriate framework for their data, balancing openness with protection of sensitive information.

## Address Ethical Considerations

As the uses of open data expand in light of developments in the generative AI ecosystem, it is more important than ever before to take action to protect data subjects and prevent harm. In order to implement open data in an ethical and a responsible way, the following actions become critical:

A. **Developing Comprehensive Ethical Guidelines:**
   a. Create detailed ethical guidelines for the collection, sharing, and use of open data, addressing emerging issues such as social license and the impact of synthetic data on privacy. This might include adapting and collectively applying existing ethical standards.

b. Establish ethical review boards or committees to oversee the implementation of these guidelines in open data projects involving new technologies such as generative AI.

**B. Implementing Advanced Privacy-Preserving Techniques:**

a. Invest in research and development of cutting-edge privacy-preserving technologies, such as differential privacy, synthetic data and federated learning, to ensure there is proper de-identification and data deletion in the context of open data for generative AI.

b. Provide resources and training to help data providers implement these technologies effectively, ensuring the privacy and security of individuals represented in open datasets being inputted into generative AI.

# APPENDIX

## Appendix 1. Open Data Action Labs Participants

Thank you to everyone who participated in our Action Labs on Open Data and Generative AI:

Alek Tarkowski, Aleš Veršič, Ashley Farley, Barbara Ubaldi, Brian Quistorff, Catherine Vogel, Cornelia Kutterer, Elena A. Kalogeropoulos, Giri Prakash, Harold Booth, Kevin Li, Kristina Podnar, Lane Dilg, Michael Tjalve, Oliver Wise, Otávio Moreira de Castro Neves, Pieter de Leenheer, Renato Berrino Malaccorto, Robert McLellan, Robin Nowok, Scott Beliveau, Shayne Longpre, Soenke Ziesche, Susan Ariel Aaronson, Tim Davies, Victoria Houed and William Hoffman.

## Appendix 2. Opportunities and Challenges of Generative AI

### Opportunities and Benefits of Generative AI

Generative AI can be leveraged to achieve positive outcomes in numerous ways. Some of the opportunities around and benefits of generative AI include:

A. **Diverse Applications:** Since its rapid and widespread adoption first began in 2022, generative AI tools have erupted in the ecosystem, making waves across numerous, diverse fields ranging from the healthcare[144] and beauty industries[145] to the worlds of art and design,[146] and even the food sector to name a few.[147] Its adaptability and transformative abilities help to increase its impact across industries and sectors.

B. **Improved Access to Information:** With the rise of generative AI-powered search engines, as well as prompt engineering and insight generation chat services, like ChatGPT, Llama and Gemini, it has become easier and faster to access knowledge at scale. The knowledge is now processed by these tools into easily digestible responses to specific questions, further reducing the time and effort required to acquire information.

C. **Increased Capabilities:** In addition to better access to knowledge, generative AI tools and chat services are also bridging skills gaps and increasing capabilities of workers and organizations by taking on certain skills-heavy tasks.[148] This can be seen with the rise of generative AI-driven coding tools, such as GitHub's Copilot for example.[149]

D. **Value Generation:** The economic benefits of generative AI are immense. In the public sector, experts estimate that generative AI could drive an impact of nearly $1.75 trillion by 2023.[150] The benefits to the private sector are even greater, with the estimated impact on the global banking industry alone amounting to an additional $200 to $340 billion in value per year.[151]

## Challenges and Risks of Generative AI

Alongside the many benefits and opportunities around generative AI, there are also a number of emerging challenges and risks associated with this technology. Some of the risks outlined in recent media include:[152]

A. **Hallucinations and Low Quality Outputs:** Generative AI models are trained to hallucinate outputs based on training data. However, sometimes the model creates inaccurate or invented outputs. This can happen when the model is not trained on appropriate data to respond to a prompt or due to underlying challenges in the algorithm.[153] In our current ecosystem, experts estimate that we will run out of high-quality training data by 2026.[154] These low quality outputs lead to an increase the risk of mis-/dis-information (including deep fakes), as well as the creation of poor quality data for future models.

---

[148] Maryam Alavi and George Westerman, "How Generative AI Will Transform Knowledge Work", *Harvard Business Review*, November 7, 2023, https://hbr.org/2023/11/how-generative-ai-will-transform-knowledge-work.

[149] "GitHub Copilot", GitHub, https://github.com/features/copilot.

[150] Miguel Carrasco et al., "Generative AI for the Public Sector: From Opportunities to Value", BCG, November 30, 2023, https://www.bcg.com/publications/2023/unlocking-genai-opportunities-in-the-government.

[151] Michael Chui et al., "The economic potential of generative AI: The next productivity frontier", McKinsey & Company, June 14, 2023, https://www.mckinsey.com/capabilities/mckinsey-digital/our-insights/the-economic-potential-of-generative-ai-the-next-productivity-frontier#business-and-society.

[152] This section does not include all challenges and risks associated with generative AI. It aims to highlight some of the common challenges discussed in recent media and is not comprehensive.

[153] "What Are AI Hallucinations?," IBM, https://www.ibm.com/topics/ai-hallucinations.

[154] Rita Matulionyte, "Researchers warn we could run out of data to train AI by 2026. What then?", *The Conversation*, November 27, 2023, https://theconversation.com/researchers-warn-we-could-run-out-of-data-to-train-ai-by-2026-what-then-216741.

B. **AI Black Boxes:** An AI black box refers to an AI system where the "internal workings are invisible to the user."[155] This lack of transparency raises important questions around the algorithm's design, its training data and the overarching model. It also leads to ethical questions related to inherent biases, data provenance and sourcing challenges, and accountability to name a few.

C. **Global Inequalities:** In many cases, generative AI technologies are exacerbating existing global socioeconomic inequalities.[156] In the workplace, for example, while generative AI tools complement (or augment) certain roles and responsibilities, they are also rendering others redundant.[157] Also, AI already consumes large amounts of energy and the rise of generative AI applications require even more water and energy–risking further exploitation of developing countries.[158]

Lastly, the design of current generative AI models reflect long standing inequalities when it comes to representation in training data. For instance, of the 7,000 languages spoken across the world, the majority of the Internet and training data available for generative AI is predominantly in English, limiting accessibility and usability of these systems in non-English contexts.[159] Open data in more languages has a key role in making the benefits of generative AI more equitably accessible.

D. **Governance Challenges:** While generative AI systems have been rapidly implemented, effective governance systems lag behind.[160] Numerous countries and intergovernmental organizations have published regulatory

---

policies and guidelines, however implementation, especially on a global scale, is challenging given the complexities of the ecosystem.[161]

E. **Security Issues:** Generative AI has the potential to open up new security risks for data users. User prompts could be incorporated in generative AI models or become a way to violate user privacy.[162] Additionally, generative AI could accelerate the unauthorized use of proprietary information or the theft of intellectual property.[163]

---

[161] Ibid.

[162] Rash, Wayne. "Generative AI Exposes Users To New Security Risks." Forbes, February 7, 2024. https://www.forbes.com/sites/waynerash/2024/02/07/generative-ai-exposes-users-to-new-security-risks/.

[163] Appel, Gil, Juliana Neelbauer, and David A. Schweidel. "Generative AI Has an Intellectual Property Problem." *Harvard Business Review*, April 7, 2023. https://hbr.org/2023/04/generative-ai-has-an-intellectual-property-problem.

# opendatapolicylab.org

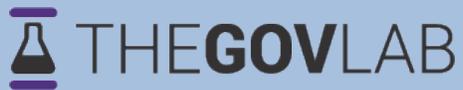 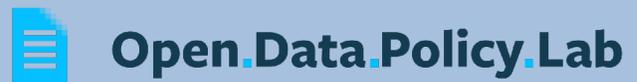